%% file: main.tex
\pdfoutput=1
\PassOptionsToPackage{table,dvipsnames}{xcolor}
\documentclass[10pt,twocolumn,letterpaper]{article}
\usepackage{iccv}

\input{preamble}

\definecolor{iccvblue}{rgb}{0.21,0.49,0.74}
\definecolor{acolor}{RGB}{0,102,255}
\definecolor{rcolor}{RGB}{255,0,0}

\usepackage[pagebackref,breaklinks,colorlinks,allcolors=iccvblue]{hyperref}
\usepackage{multirow}
\title{
\raisebox{-0.3\height}{
\begin{tikzpicture}
    \clip (0.55,0.55) circle(0.55);
    \node at (0.55,0.55) {\includegraphics[width=0.03\textwidth]{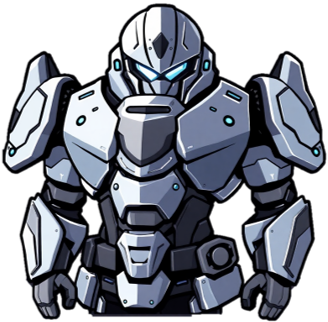}}; 
\end{tikzpicture}
}
\hspace{-0.5cm}
ARMOR: Empowering Multimodal Understanding Model with Interleaved Multimodal Generation Capability
}

\author{
  \textbf{Jianwen Sun}$^{1,2}$\thanks{Equal contribution} \quad
  \textbf{Yukang Feng}$^{1,2}$\footnotemark[1] \quad
  \textbf{Chuanhao Li}$^{5}$ \quad
  \textbf{Fanrui Zhang}$^{2,3}$ \quad
  \textbf{Zizhen Li}$^{1,2}$ \quad \\
  \textbf{Jiaxin Ai}$^{2,4}$ \quad
  \textbf{Sizhuo Zhou}$^{2,3}$ \quad
  \textbf{Yu Dai}$^{1}$ \quad
  \textbf{Shenglin Zhang}$^{1}$ \quad \\
  \textbf{kaipeng Zhang}$^{2,5}$\thanks{Corresponding author.} \quad \\
  Nankai University$^{1}$\quad Shanghai Innovation Institute$^{2}$\quad University of Science and Technology of China$^{3}$ \\ Wuhan University $^{4}$\quad Shanghai AI Laboratory$^{5}$ \\
  \url{https://github.com/finyorko/armor} \quad \\
  \textcolor{red}{This paper is a refined version of the previous ARMOR.}
}

\begin{document}
\maketitle
\input{sec/0_abstract}

\input{sec/1_intro}
\input{sec/2_related_work}
\input{sec/3_method}

\input{sec/4_exp}
\input{sec/5_conclusion}
{
    \small
    \bibliographystyle{ieeenat_fullname}
    \bibliography{main}
}

\input{sec/X_suppl}
\end{document}

%% file: preamble.tex
\usepackage{tabularx}
\usepackage[utf8]{inputenc}
\usepackage{tcolorbox}
\usepackage{lipsum}
\usepackage{graphicx}
\usepackage{float}
\usepackage{amsmath}
\usepackage{changepage}
\usepackage{placeins}
\usepackage{svg}
\usepackage{amssymb}   
\usepackage{makecell} 
\usepackage{colortbl}

\usepackage{tikz}
\usetikzlibrary{shapes,arrows,positioning}

\usepackage{enumitem}
\usepackage{subcaption}

\usepackage{caption}
\usepackage{mdframed}

\usepackage[table]{xcolor}

%% file: sec/0_abstract.tex
\begin{abstract}

Unified multimodal understanding and generation have recently received much attention in the area of vision and language.
Existing UniMs are designed to simultaneously learn both multimodal understanding and generation capabilities, demanding substantial computational resources, and often struggle to generate interleaved text-image.
We present ARMOR, a resource-efficient and pure autoregressive framework that achieves both understanding and generation by fine-tuning existing multimodal large language models (MLLMs).
Specifically, ARMOR extends existing MLLMs from three perspectives:
(1) For model architecture, an asymmetric encoder-decoder architecture with a forward-switching mechanism is introduced to unify embedding space integrating textual and visual modalities for enabling natural text-image interleaved generation with minimal computational overhead.
(2) For training data, a meticulously curated, high-quality interleaved dataset is collected for fine-tuning MLLMs.
(3) For the training algorithm, we propose a ``what or how to generate'' algorithm to empower existing MLLMs with multimodal generation capabilities while preserving their multimodal understanding capabilities, through three progressive training stages based on the collected dataset.
Experimental results demonstrate that ARMOR upgrades existing MLLMs to UniMs with promising image generation capabilities, using limited training resources.
Our code will be released soon at \url{https://github.com/finyorko/armor}.
\end{abstract}

%% file: sec/1_intro.tex
\section{Introduction}
\label{sec:intro}

Unified understanding and generation is a crucial direction in the development of vision-and-language models, requiring a model to simultaneously handle understanding tasks (\textit{e.g.}, visual question answering) and generation tasks (\textit{e.g.}, text-to-image generation).
Existing Unified Models (UniMs) for understanding and generation, such as Show-o \cite{showo} and Janus-pro \cite{januspro},
are designed to simultaneously learn both multimodal understanding and generation capabilities.
Despite their impressive performance, the training of such models demands substantial computational resources, which significantly hinders their scalability and ability to accommodate personalized modifications.
Furthermore, most UniMs struggle to generate interleaved text-image.
To this end, we present a resource-efficient autoregressive framework named ARMOR, which fine-tunes existing multimodal large language model (MLLMs) to upgrade them to UniMs from three perspectives, including model architecture, training data and training algorithm.

\begin{figure*}[t]
  \centering
   \includegraphics[width=1.0\linewidth]{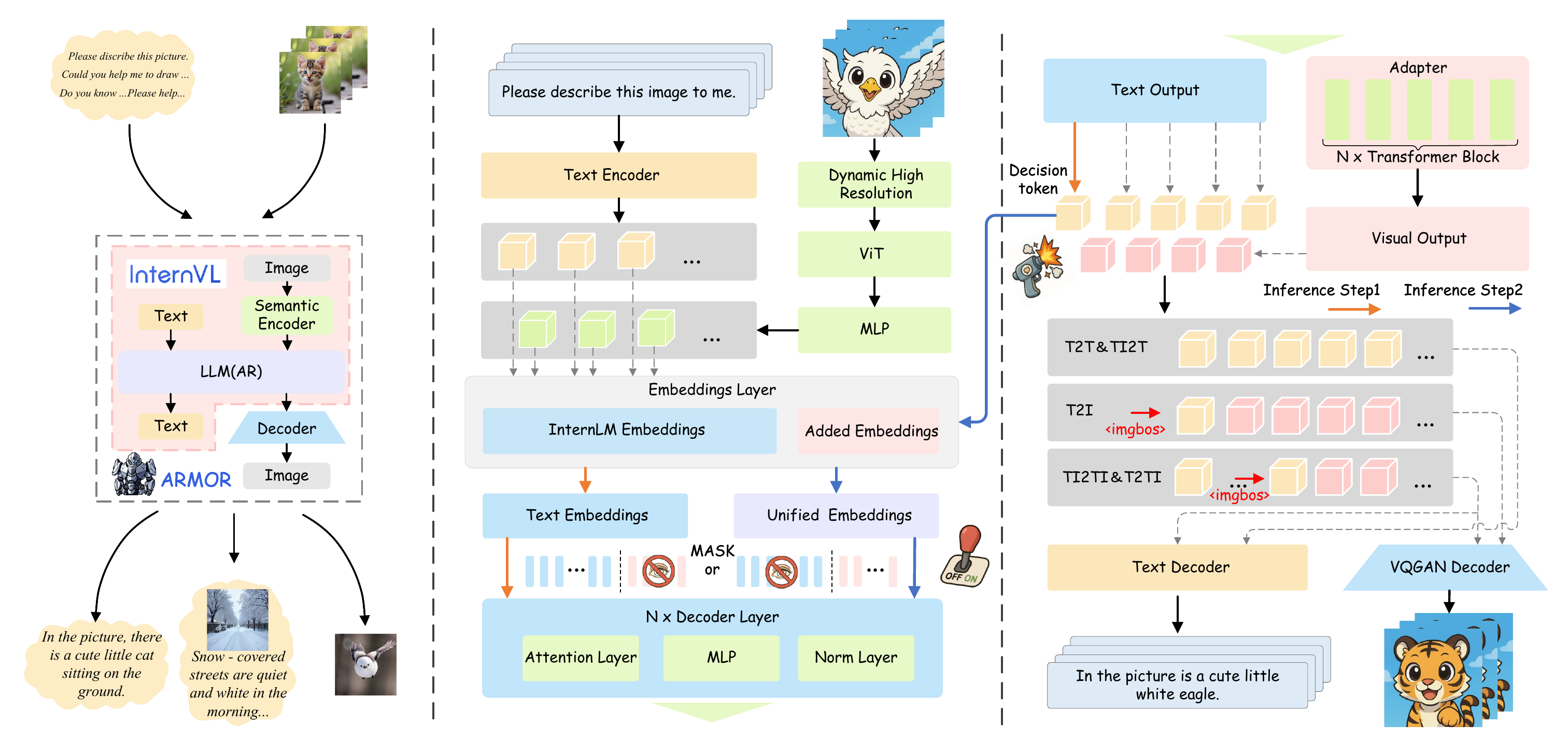}
   \vspace{-15pt}
   \caption{Schematic diagram of the proposed ARMOR framework. This framework consists of an MLLM and a pre-trained VQVAE decoder; The codebook of MLLM is expanded to accommodate image information. Generation is completely based on the autoregressive architecture, and special tokens are used as switches for modality switching.}

   \label{fig:architecture}
\end{figure*}

We first introduce an asymmetric encoder-decoder architecture into existing MLLMs to enable them output natural interleaved text-image.
Specifically, ARMOR fully retains the encoder and decoder in MLLMs while incorporating an asymmetric image decoder, as shown in Figure ~\ref{fig:architecture}.
In doing so, the strong semantic perception and understanding capabilities of the MLLMs are almost preserved, while the asymmetric image decoder enables image generation with minimal computational overhead.
To alleviate the long-tail distribution problem across different modalities in the answer space, we propose a forward-switching mechanism to dynamically control which modality's answer space is used for prediction based on the model input.

Then we collect a meticulously curated, high-quality interleaved dataset, and propose a ``What or How to Generate" (WoHG) training algorithm to fune-tune existing MLLMs with the collected dataset.
The training algorithm consists of three training stages with different objectives: \textit{what to generate}, \textit{how to generate} and \textit{how to answer better}.
These different objectives are achieved by freezing different parameters and training on different types of data.
In the first stage, MLLMs learn to decide the response modality, such as generating text or producing images.
In the second stage, MLLMs specifically improve their shortcomings—image generation capability, further enhancing the quality of the generated images.
In the third stage, MLLMs refine their responses to better integrate text and visual modalities, providing high-quality text-image interleaved responses.
The three-stage training algorithm ensures structured progression by first establishing modality awareness, then targeting capability gaps, and ultimately achieving holistic multimodal synergy through iterative refinement.

To validate the effectiveness of the proposed ARMOR framework, we incorporate InternVL2.5 \cite{internvl2.5} into ARMOR (named ``Ours'' here), and conduct extensive experiments on 9 benchmarks to evaluate multimodal understanding and generation capabilities.
Experimental results demonstrate that Ours outperforms existing UniMs in multimodal understanding with a large margin (\textit{e.g.}, 78.5 and 62.6 in score for Ours and Janus-pro \cite{januspro} on the MMB benchmark \cite{mmbench}, respectively)
while achieving comparable performance for multimodal generation (\textit{e.g.}, 0.51 and 0.39 in score for Ours and Chameleon \cite{chameleon} on the GenEval benchmark \cite{geneval}, respectively).
Notably, ARMOR only introduces $\sim$10\% more parameters for fine-tuning InternVL2.5, whereas existing UniMs require full parameter training from scratch.

Our contributions are summarized as follows.
(1) We propose the first framework to build UniMs by empowering existing MLLMs with generation capabilities in a resource-efficient manner.
(2) We present an asymmetric encoder-decoder architecture to enable natural text-image interleaved generation for MLLMs while minimizing computational overhead.
(3) We collect a high-quality dataset for training MLLMs, and propose a WoHG training algorithm to upgrade existing MLLMs to UniMs while preserving their understanding capabilities through three progressive training stages on the collected dataset.
(4) Extensive experimental results on 9 benchmarks demonstrate the effectiveness of our framework, further affirming the potential of a fully autoregressive architecture for building UniMs.

%% file: sec/2_related_work.tex
\section{Related Work}
\label{sec:related_work}

\subsection{Multimodal Understanding}

CLIP~\cite{clip} pioneers cross-modal alignment via contrastive learning, inspiring MLLMs that bridge vision encoders and large language models.
Two dominant alignment strategies have emerged:
1) Explicit attention interaction, exemplified by BLIP-2's Q-Former~\cite{blip2} and Flamingo’s cross-attention~\citep{flamingo}, enables deep vision-language fusion by projecting visual features into text-aligned tokens.
2) Implicit space mapping, as seen in InternVL~\citep{chen2024internvl} and Qwen2.5-VL~\citep{Qwen2.5-VL}, transforms visual features into token sequences using MLPs for modality alignment.
While differing in encoder choices, alignment designs, and training data strategies, most MLLMs share core components (\textit{e.g.}, pretrained vision backbones, lightweight adapters, and autoregressive text decoders),
and continue to evolve through improved visual representations and more efficient alignment methods.

\subsection{Visual Generation}

The field of visual generation has evolved through autoregressive and diffusion-based approaches.
Early autoregressive models~\citep{parmar2018image, taming, ravuri2019classification, chen2020generative} leverage Transformer for pixel-level text-to-image synthesis but face challenges in high-resolution fidelity.
Subsequent token-based methods~\citep{llamagen} improve sequence prediction but remained limited in fine-grained detail.
In contrast, diffusion models like stable diffusion series~\citep{diffusion, sdxl, sd3}, DALL-E 2~\citep{dalle2}, and FLUX~\citep{flux2024} gain prominence by iteratively denoising images to achieve state-of-the-art visual quality.
These methods excel in generating high-fidelity, detailed outputs for text-to-image tasks.
However, diffusion frameworks focus primarily on generation, lacking integrated text-image understanding capabilities.
While autoregressive models emphasize cross-modal dependencies, diffusion approaches prioritize photorealism through progressive refinement.

\subsection{Unified Understanding and Generation}

Recently, an increasing number of studies focus on unified models for understanding and generation.
Next-GPT~\citep{next-gpt} and SEED-X~\citep{seed-x} achieve this goal by combining separate understanding and generation systems.
Show-o~\citep{showo} and TransFusion~\citep{transfusion} employ hybrid diffusion-autoregressive methods.
Emu2~\citep{emu2} use a fully autoregressive architecture for predicting the next multimodal elements, using classification tokens for text prediction and regressing visual embeddings for image prediction.
Chameleon~\cite{chameleon}, VILA-U~\citep{vilau}, and Emu3~\citep{vilau} convert images into tokens, intertwining image tokens with text tokens from the very beginning, to support joint text-image reasoning and autoregressive prediction.
Although the above models are unified for understanding and generation,
they are trained to simultaneously learn both multimodal understanding and generation capabilities,
which is resource-intensive and leads to a failure in outputting interleaved text-image.
In contrast, we propose the ARMOR framework, which builds UniMs by fine-tuning existing MLLMs in a resource-efficient manner to empower them with the capabilities text-to-image generation and outputting interleaved text-image.

%% file: sec/3_method.tex
\section{ARMOR Framework}

\subsection{Preliminary}

To improve training performance, we introduce a weighted loss calculation method, allowing for dynamic adjustments at different training stages.
We optimize the loss by adopting two learning objectives: the loss function for text prediction and for image prediction. We create label masks to ensure that the model can separately compute the loss of text or image prediction.

\noindent \textbf{Text Prediction Loss Calculation:}
\begin{align}
    \mathcal{L}_{\text{text}} &= - \sum_{t=1}^{T} \mathbb{I}_{\text{text}}(t) \cdot \log P_1(y_t \mid y_{<t}, M)
\end{align}
\begin{itemize}
    \item \(T\): total length of the target sequence (total time steps)
    \item \(\mathbb{I}_{\text{text}}(t)\): 1 if time step \(t\) is in text output mode
    \item \(y_t\): the \(t\)-th token in the target sequence (text token)
    \item \(y_{<t}\): the previously generated tokens before time step \(t\)
    \item \(M\): the joint representation of the multimodal input
    \item \(P_1\): probability distribution for text output
\end{itemize}

\noindent \textbf{Image Prediction Loss Calculation:}
\begin{align}
    \mathcal{L}_{\text{img}} &= - \sum_{t=1}^{T} \mathbb{I}_{\text{img}}(t) \cdot \log P_2(y_t \mid y_{<t}, M)
\end{align}
\begin{itemize}
    \item \(\mathbb{I}_{\text{img}}(t)\): 1 if time step \(t\) is in image output mode
    \item \(y_t\): the \(t\)-th token in the target sequence (image token)
    \item \(P_2\): probability distribution for image output
\end{itemize}

\noindent \textbf{Overall Loss Calculation:}
\begin{align}
    \mathcal{L}_{\text{total}} &= \alpha \mathcal{L}_{\text{text}} + \beta \mathcal{L}_{\text{img}}
\end{align}
\begin{itemize}
    \item \(\alpha\): weight for text loss (default=1.0)
    \item \(\beta\): weight for image loss (default=1.0)
\end{itemize}

\subsection{Architecture}

Rather than focusing on enhancing the comprehension capabilities of UniMs, we have chosen to explore the generative potential of MLLMs. So the core challenge we face is how to enable UniMs to achieve generation capability comparable to those of MLLMs. Therefore, we have to solve the following two problems:
\begin{enumerate} 
    \item How can MLLMs obtain generation ability? 
    \item While MLLMs acquire generation ability, how can we avoid catastrophic forgetting of its understanding abilities?
\end{enumerate}

In the question of how to endow MLLM with generation ability, we choose to integrate a pre-trained VQGAN~\citep{vqgan}-decoder (originating from Chameleon) for InternVL2.5~\citep{internvl2.5} (vit-mlp-llm) and achieve a breakthrough in generation ability whit an asymmetric encoder-decoder manner. Specifically, we adopted a simple and effective approach: integrating text information and image information by extending the indices of the pre-trained VQGAN to codebook of MLLM.

Specifically, we added new tokens from Table. ~\ref{tab:tokens} to the InternVL2.5 model. Some of these tokens are used to map the indices of VQGAN, and we expanded other related structures of the model (embedding layer and output layer) to enable the model to learn image information. For the newly added parameters, we use random initialization and freeze the weights of the original text tokens.

Regarding the problem of how to inherit the original capabilities and avoid catastrophic forgetting during the training process, we proposed two schemes in the early stage of the research:
\begin{enumerate}
    \item Use data hybrid training. Collect a large amount of training data for MLLM or distill high-quality Supervised Fine-Tuning (SFT) data. Mix text-image interleaved data to train the model backbone and retain the original capabilities as much as possible during the training process.
    \item Modify the model structure. Add additional modules to enable the model's original weights to participate in training as little as possible, thereby reducing the impact on the model's original capabilities.
\end{enumerate}

After comparison and discussion, the first scheme requires training a large number of the model's original parameters. Moreover, the ability level after training is related to the quality of SFT data, making it difficult to guarantee the same level as before. In addition, it also needs to deal with the heavy workload of data distillation and fusion. Therefore, we finally decided to adopt the second scheme.
We added several transformer layers as a adapter for the backbone. The newly added transformer layers are consistent with the model. Such changes will not damage the original structure of the model. The unification of understanding and generation capabilities is achieved entirely based on the next token prediction method.

\begin{figure}[t]
  \centering
   \includegraphics[width=1\linewidth]{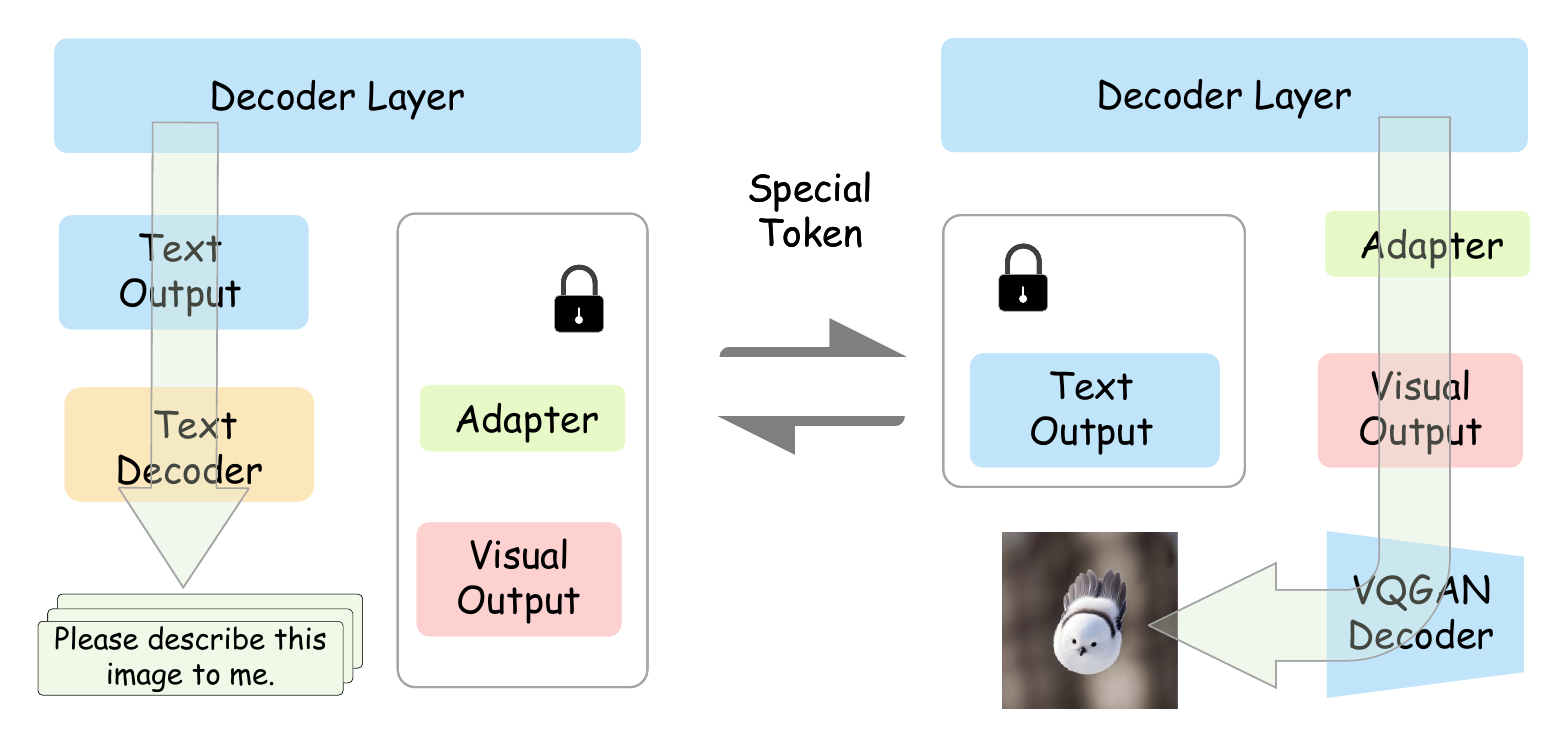}
   \caption{Forward switching mechanism.}
   \label{fig:forward}
\end{figure}

After modifying the model according to the above ideas, when predicting image tokens in the forward process, the output layer needs to conduct a full-scale classification of the entire codebook. A large number of text tokens lacking image information may interfere with the prediction of image tokens. Conversely, the same is true. This redundant classification increases the learning difficulty of the model. Therefore, we designed forward-switching mechanism, which outputs through two different output heads according to different modalities. The architecture of the model is shown in Figure ~\ref{fig:architecture}. The forward-switching is controlled by special tokens and the mechanism of forward-switching is shown in Figure ~\ref{fig:forward}, and the relationship between the newly added tokens and the output layer is described in Table ~\ref{tab:tokens}. In the process of autoregressive output, there is no need to classify all tokens in the codebook. Instead, only focus on the information of a certain modality. This gated classification output mechanism not only reduces the model's search space but also enables the model to autonomously determine the modal generation path and achieve natural mixed-modality output. Furthermore, all the modifications we made to the model only added a total of \textbf{0.9 billion} trainable parameters.

\begin{table*}[ht]
\centering
\begin{tabular}{|l|c|l|}
\hline
\textbf{Token Type} & \textbf{Head} & \textbf{Functional Description} \\
\hline
\multicolumn{3}{|l|}{\textit{Special Tokens:}} \\
\hline
\texttt{<imgbos>} & text output & Switch model to visual output mode, begin image generation \\
\texttt{<imgend>} & visual output & Terminate image generation, revert to text output mode \\
\texttt{<imgpad>} & visual output & Padding placeholder in image token sequences \\
\hline
\multicolumn{3}{|l|}{\textit{Image Content Tokens:}} \\
\hline
8192 image tokens & visual output & Content representation tokens \\
\hline
\end{tabular}
\caption{Special token and image content token specifications.}
\label{tab:tokens}
\end{table*}

\begin{figure*}[t]
  \centering
  \resizebox{0.75\textwidth}{!}{
  \begin{subfigure}{0.25\linewidth}
    \includegraphics[width=\linewidth]{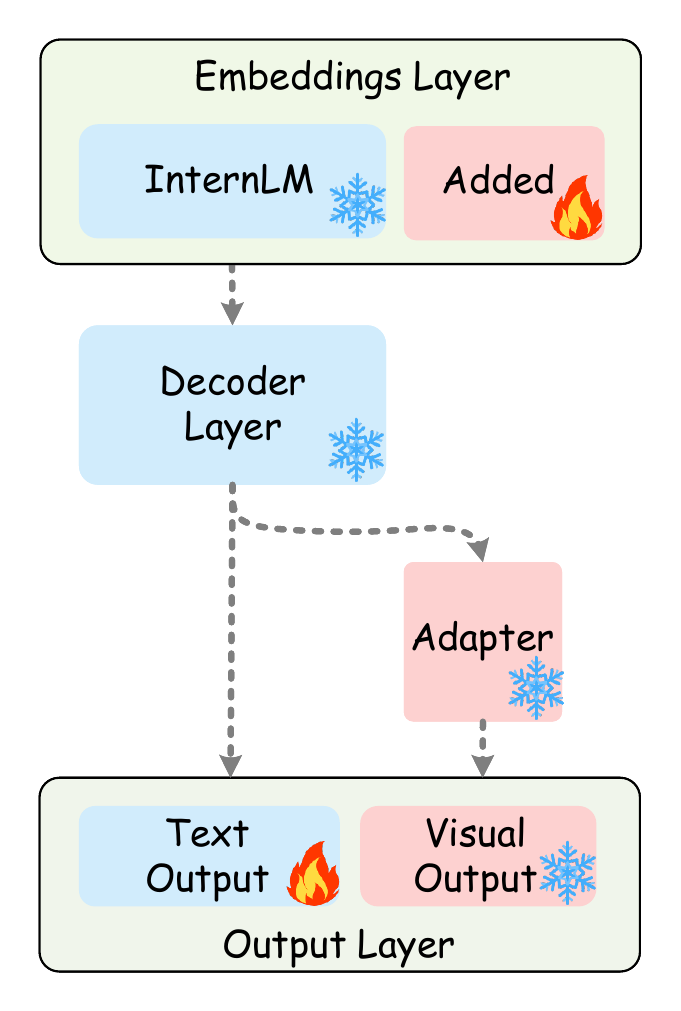}
    \caption{step 1.}
    \label{fig:step1}
  \end{subfigure}  
  \hspace{0.05\linewidth}
  \begin{subfigure}{0.25\linewidth}
    \includegraphics[width=\linewidth]{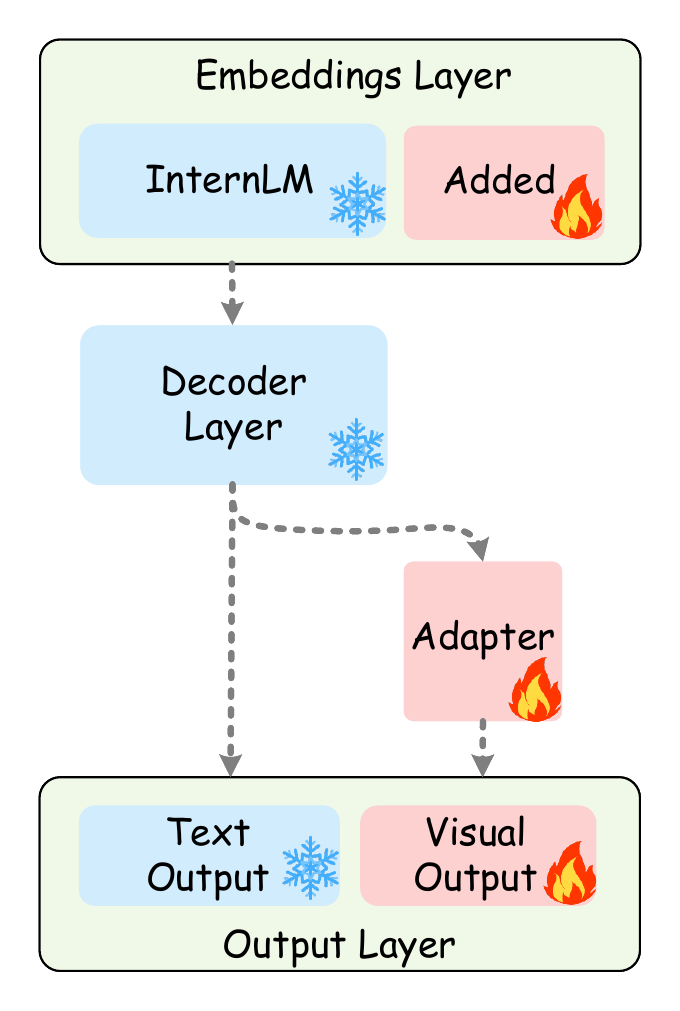}
    \caption{step 2.}
    \label{fig:step2}
  \end{subfigure}
  \hspace{0.05\linewidth}
  \begin{subfigure}{0.25\linewidth}
    \includegraphics[width=\linewidth]{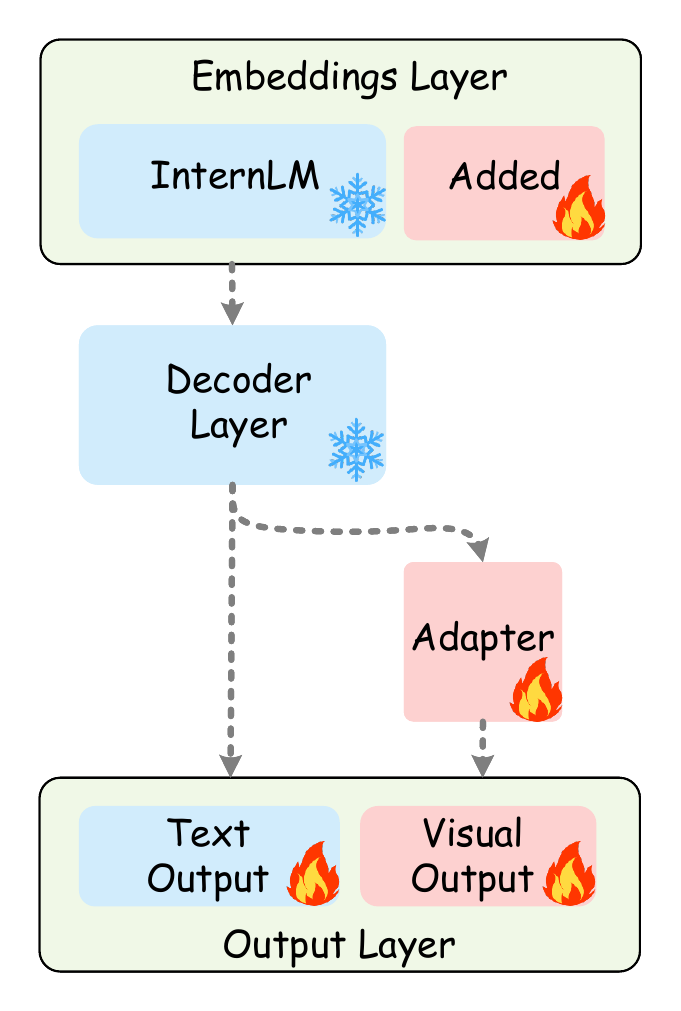}
    \caption{step 3.}
    \label{fig:step3}
  \end{subfigure}
  }
    \caption{Demonstration of the proposed three-stage WoHG training algorithm.}
    \label{fig:step_all}
\end{figure*}

\subsection{Data Collection}
To achieve the model's mixed modality input and output capabilities, we consider the data required for training through the following several scenarios:

\begin{description}[leftmargin=2em,style=sameline]
    \item[Q1:] When should the model respond solely with text? 
    \begin{itemize}
        \item \textbf{Applicable scenarios:} Natural language dialogue tasks and multimodal understanding tasks requiring textual responses without image generation
        \item \textbf{Training data:} \texttt{t2t} (Text-to-Text) and \texttt{ti2t} (Text-Image-to-Text) datasets
    \end{itemize}
    \item[Q2:] When should the model only generate images?
    \begin{itemize}
        \item \textbf{Applicable scenarios:} Explicit image generation tasks with direct visual content requests
        \item \textbf{Training data:} \texttt{t2i} (Text-to-Image) datasets
    \end{itemize}
    \item[Q3:] When should the model produce mixed responses?
    \begin{itemize}
        \item \textbf{Applicable scenarios:} Multimodal interaction tasks mirroring human communication patterns requiring combined textual and visual outputs
        \item \textbf{Training data:} \texttt{t2ti} (Text-to-Text-Image) datasets
    \end{itemize}
\end{description}

To sum up, training data consists of the following types:

\begin{enumerate}
    \item \textbf{t2t}: Standard dialogue-based questions and answers. This part of the dataset is entirely self-constructed.
    \item \textbf{ti2t}: Image comprehension tasks where the model is given an image and asked to generate textual descriptions. This part of the dataset is selected from ShareGPT4V~\citep{sharegpt4v}, LLaVA~\cite{liu2023improvedllava} and some data distilled from Internvl2.5~\cite{internvl2.5}.
    \item \textbf{t2i}: Text-based image generation tasks, where the model generates an image based on a given prompt. This part of the dataset is composed of three parts: self-construction, screening from LAION-nolang-aesthetics-27M~\citep{laion}, and screening from text-to-image-2M ~\citep{zk_2024}.          
    \item \textbf{t2ti}: Mixed-modality dialogue tasks where both textual responses and generated images are required. This part of the dataset is entirely self-constructed.
\end{enumerate}

\subsection{WoHG Training Algorithm}
\label{sec:training}

For the model training, we proposed a three-stage training algorithm named WoHG (what or how to generate). We designed specific training objectives for each stage to ensure that the model's capabilities steadily improve during the training process without causing the problem of catastrophic forgetting. The specific modules trained in each stage are shown in Figure ~\ref{fig:step_all}.

\noindent \textbf{(a) First Stage: What to Generate?}

We deem that it is important for the model to autonomously make different responses according to the form of the question. Relying on external information to select a response mode will affect the interaction. Therefore, what content to generate in a conversational context is the training focus of this step. We call this ``what to generate".

In this stage, the types and quantities of datasets we employ are t2t (100K), ti2t (300K), t2i (100K) and t2ti (100K).
We adjust the calculation weights of loss function as follows: $\alpha = 1.0 $ and $\beta = 0.0$.
The trainable parameters in this stage are shown in Figure ~\ref{fig:step1}.

In the first stage of training, we use the datasets of the above four different question types to train the model's ability to distinguish question types. This stage lays the foundation for the model’s ability to handle multimodal inputs and outputs in later stages.

\noindent \textbf{(b) Second Stage: How to Generate?}

After the first stage of training, the model is able to generate appropriate answer pattern in any given question. The second stage requires activating the model's image generation ability and the corresponding relationship between images and text content. Train all parameters related to generation of the model to ensure that it can accurately generate images that meet the requirements according to the input text information. So the core of the second stage is how to generate appropriate images when facing generation demands. We call this ``how to generate".

In this stage, the types and quantities of datasets are t2i (2.5M) and t2ti (2.5M).
We adjust the calculation weights of loss function as follows: $\alpha = 0.0 $ and $\beta = 1.0$.
The trainable parameters in this stage are shown in Figure ~\ref{fig:step2}.
After this stage of training, the model is able to generate impressive images. The changes in image generation quality during part of the training process are shown in Figure ~\ref{fig:show-fig}.

\noindent \textbf{(c) Third Stage: How to Answer Better?}

In this stage, we use a carefully selected high-quality text-image interleaved dataset to fine-tune the model. The focus of this stage is how to generate better text-image interleaved responses and ensure a better synergistic effect between the generated text and images. We call this ``how to answer better".

In this stage, the types and quantities of datasets are t2t (50K), ti2t (300K), t2i (300K) and t2ti (50K).
Furthermore, we adjust the calculation weights of the loss function as follows:$\alpha = 1.0 $ and $\beta = 1.0$.

Through this stage of training, the model is able to output high-quality interleaved messages. It is worth mentioning that for the ti2ti dialogue pattern, we did not specifically integrate this ability for the model. However, due to the characteristics of ARMOR, if there is a demand for image generation in the question, then ARMOR can naturally predict relevant image information based on the content of the text answer. Examples are recorded in supplementary materials. The trainable parameters of this stage are shown in Figure ~\ref {fig:step3}. In addition, the proportion of the dataset in the training phase and the analysis can be found in supplementary materials.

During the three - stage training process of the model, we used AdamW as the optimizer, where  $ \beta $ 1 was 0.9 and $ \beta $ 2 was 0.999. In all three stages, cosine annealing was used for the learning rate schedule. The gradient clipping threshold was set to 1.0, and the weight decay was 0.05. Finally, the learning rates for the three stages were set to 4e - 5, 1e - 4, and 5e - 5 respectively. The training of all stages of the model was carried out on 8 * H100 GPUs.

MLLMs naturally support interleaved text-image input for handling comprehension tasks. After the above-mentioned training, ARMOR extends its generation ability to achieve interleaved text-image output. Ultimately, ARMOR can successfully integrate these two abilities to realize \textbf{interleaved text-image input-output}.

%% file: sec/4_exp.tex
\begin{figure*}[t]
  \centering
  \resizebox{0.98\textwidth}{!}{
    \begin{subfigure}{0.13\linewidth}
      \includegraphics[width=\linewidth]{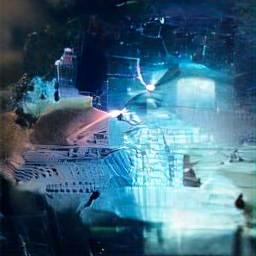}
      \label{fig:step1-1}
    \end{subfigure}
    \hspace{0.006\linewidth}
    \begin{subfigure}{0.13\linewidth}
      \includegraphics[width=\linewidth]{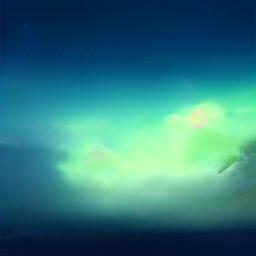}
      \label{fig:step1-2}
    \end{subfigure}
    \hspace{0.006\linewidth}
    \begin{subfigure}{0.13\linewidth}
      \includegraphics[width=\linewidth]{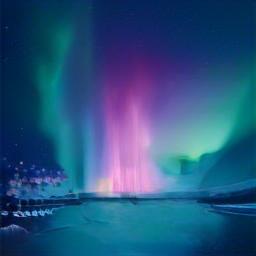}
      \label{fig:step1-3}
    \end{subfigure}
    \hspace{0.006\linewidth}
    \begin{subfigure}{0.13\linewidth}
      \includegraphics[width=\linewidth]{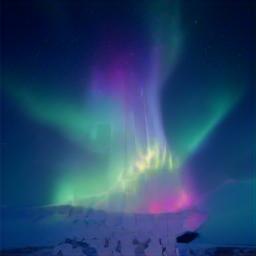}
      \label{fig:step1-4}
    \end{subfigure}
    \hspace{0.006\linewidth}
    \begin{subfigure}{0.13\linewidth}
      \includegraphics[width=\linewidth]{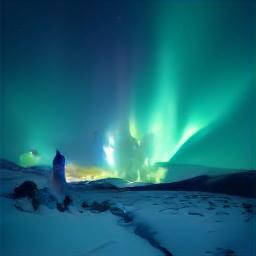}
      \label{fig:step1-5}
    \end{subfigure}
    \hspace{0.006\linewidth}
    \begin{subfigure}{0.13\linewidth}
      \includegraphics[width=\linewidth]{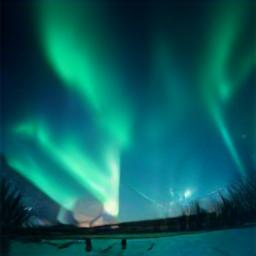}
      \label{fig:step1-6}
    \end{subfigure}
    \hspace{0.006\linewidth}
    \begin{subfigure}{0.13\linewidth}
      \includegraphics[width=\linewidth]{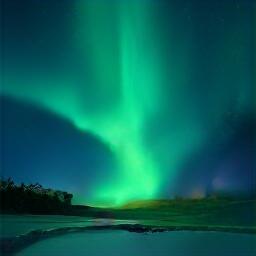}
      \label{fig:step1-7}
    \end{subfigure}
    \hspace{0.006\linewidth}
    \begin{subfigure}{0.13\linewidth}
      \includegraphics[width=\linewidth]{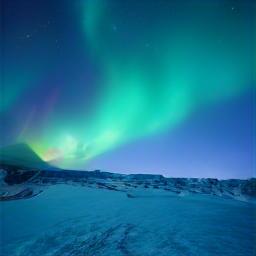}
      \label{fig:step1-8}
    \end{subfigure}
  }
  \resizebox{0.98\textwidth}{!}{
    \begin{subfigure}{0.13\linewidth}
      \includegraphics[width=\linewidth]{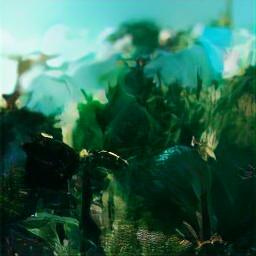}
      \label{fig:step2-1}
    \end{subfigure}
    \hspace{0.006\linewidth}
    \begin{subfigure}{0.13\linewidth}
      \includegraphics[width=\linewidth]{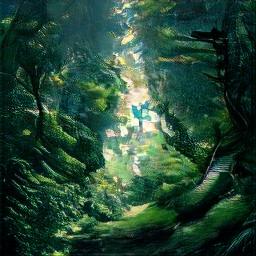}
      \label{fig:step2-2}
    \end{subfigure}
    \hspace{0.006\linewidth}
    \begin{subfigure}{0.13\linewidth}
      \includegraphics[width=\linewidth]{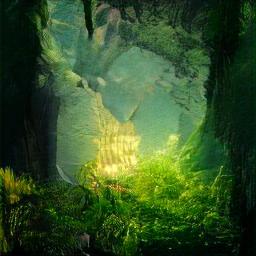}
      \label{fig:step2-3}
    \end{subfigure}
    \hspace{0.006\linewidth}
    \begin{subfigure}{0.13\linewidth}
      \includegraphics[width=\linewidth]{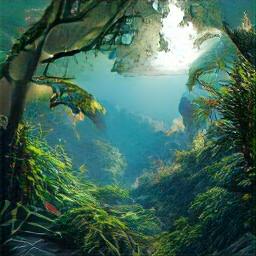}
      \label{fig:step2-4}
    \end{subfigure}
    \hspace{0.006\linewidth}
    \begin{subfigure}{0.13\linewidth}
      \includegraphics[width=\linewidth]{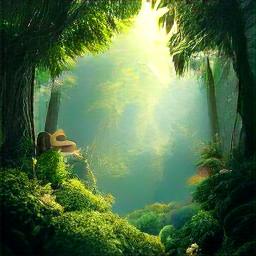}
      \label{fig:step2-5}
    \end{subfigure}
    \hspace{0.006\linewidth}
    \begin{subfigure}{0.13\linewidth}
      \includegraphics[width=\linewidth]{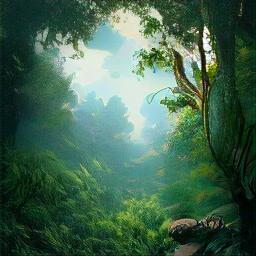}
      \label{fig:step2-6}
    \end{subfigure}
    \hspace{0.006\linewidth}
    \begin{subfigure}{0.13\linewidth}
      \includegraphics[width=\linewidth]{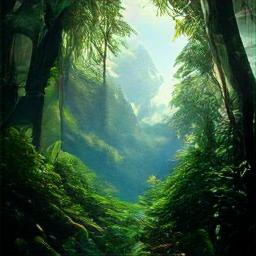}
      \label{fig:step2-7}
    \end{subfigure}
    \hspace{0.006\linewidth}
    \begin{subfigure}{0.13\linewidth}
      \includegraphics[width=\linewidth]{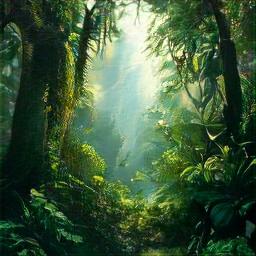}
      \label{fig:step2-8}
    \end{subfigure}
  }
  \caption{Changes in image generation quality during part of the training process (epochs: 4, 6, 8... 18, \textbf{from left to right}). Prompt 1: ``\textit{Could you generate an image of the aurora for me?}"; Prompt 2: ``\textit{Please help me draw a picture of the tropical rainforest.}".}
  \label{fig:show-fig}
\end{figure*}

\begin{figure*}[t]
  \centering
  \resizebox{0.98\textwidth}{!}{
    \begin{subfigure}{0.13\linewidth}
      \includegraphics[width=\linewidth]{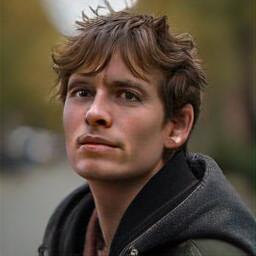}
      \vspace{1pt}
    \end{subfigure}
    \hspace{0.005\linewidth}
    \begin{subfigure}{0.13\linewidth}
      \includegraphics[width=\linewidth]{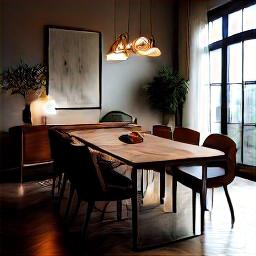}
      \vspace{1pt}
    \end{subfigure}
    \hspace{0.005\linewidth}
    \begin{subfigure}{0.13\linewidth}
      \includegraphics[width=\linewidth]{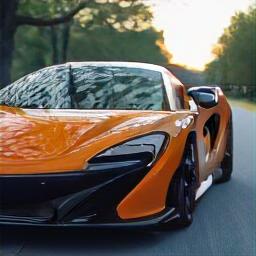}
      \vspace{1pt}
    \end{subfigure}
    \hspace{0.005\linewidth}
    \begin{subfigure}{0.13\linewidth}
      \includegraphics[width=\linewidth]{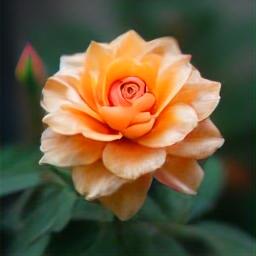}
      \vspace{1pt}
    \end{subfigure}
    \hspace{0.02\linewidth}
    \begin{subfigure}{0.13\linewidth}
      \includegraphics[width=\linewidth]{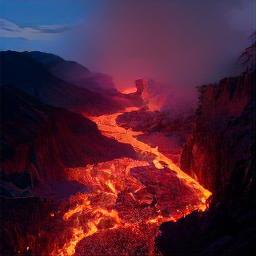}
      \vspace{1pt}
    \end{subfigure}
    \hspace{0.005\linewidth}
    \begin{subfigure}{0.13\linewidth}
      \includegraphics[width=\linewidth]{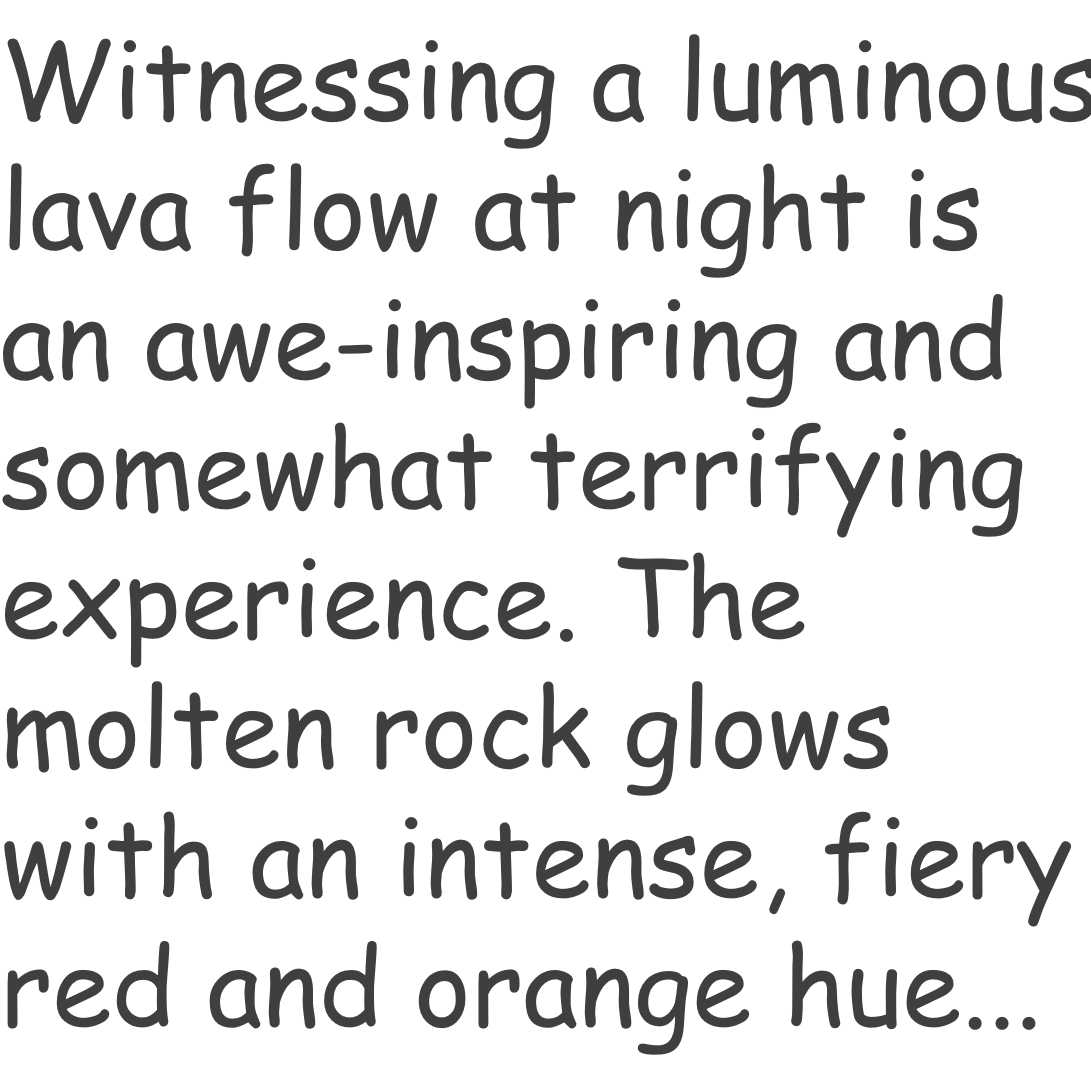}
      \vspace{1pt}
    \end{subfigure}
    \hspace{0.005\linewidth}
    \begin{subfigure}{0.13\linewidth}
      \includegraphics[width=\linewidth]{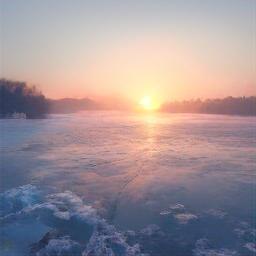}
      \vspace{1pt}
    \end{subfigure}
    \hspace{0.005\linewidth}
    \begin{subfigure}{0.13\linewidth}
      \includegraphics[width=\linewidth]{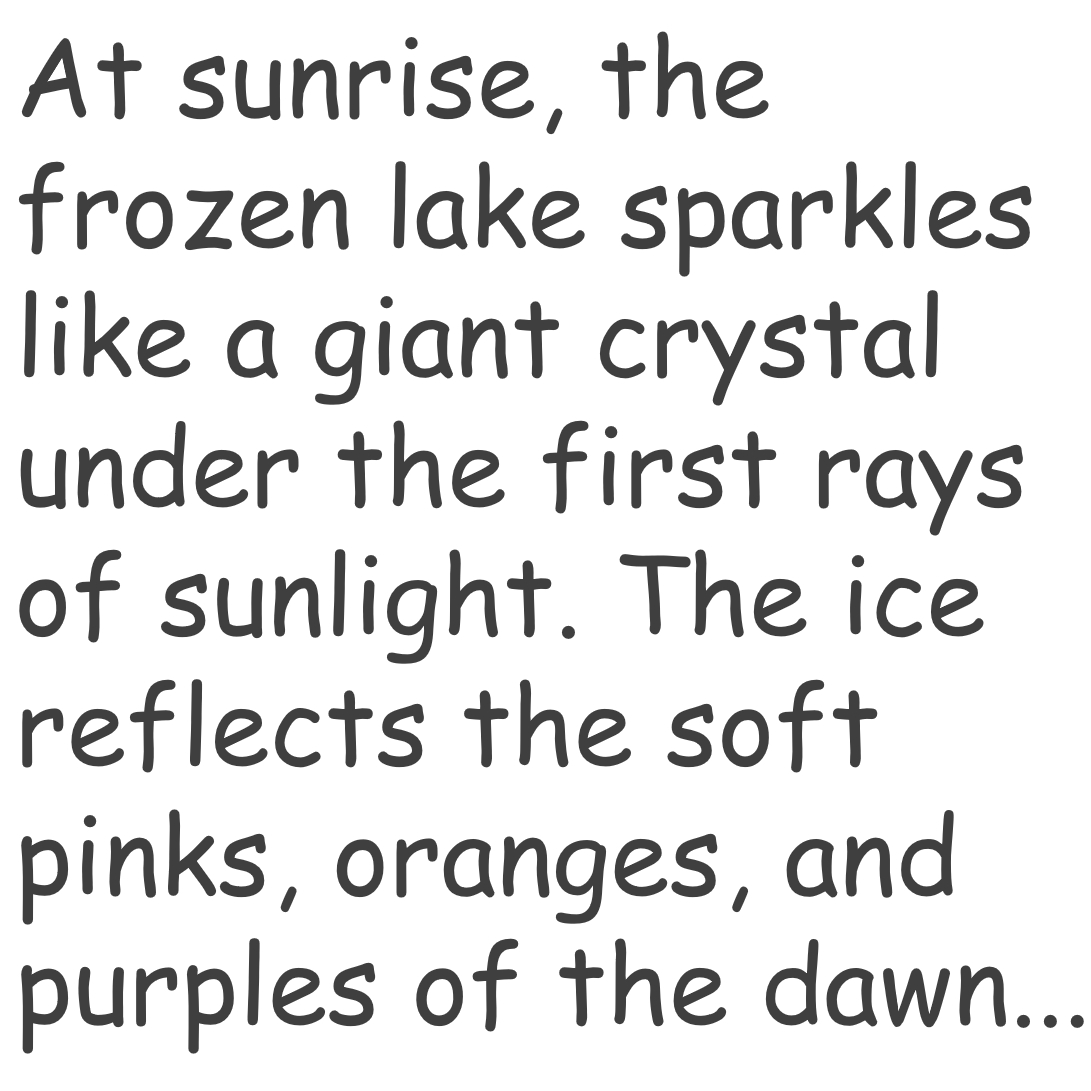}
      \vspace{1pt}
    \end{subfigure}
  }
  \resizebox{0.98\textwidth}{!}{
    \begin{subfigure}{0.13\linewidth}
      \includegraphics[width=\linewidth]{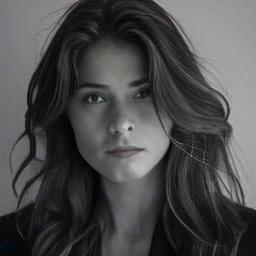}
    \end{subfigure}
    \hspace{0.005\linewidth}
    \begin{subfigure}{0.13\linewidth}
      \includegraphics[width=\linewidth]{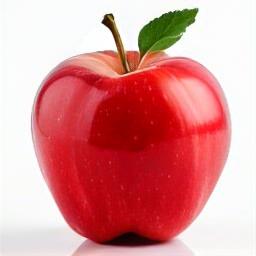}
    \end{subfigure}
    \hspace{0.005\linewidth}
    \begin{subfigure}{0.13\linewidth}
      \includegraphics[width=\linewidth]{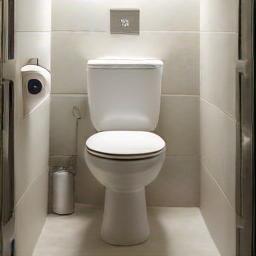}
    \end{subfigure}
    \hspace{0.005\linewidth}
    \begin{subfigure}{0.13\linewidth}
      \includegraphics[width=\linewidth]{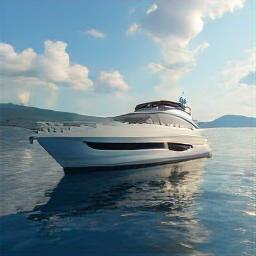}
    \end{subfigure}
    \hspace{0.02\linewidth}
    \begin{subfigure}{0.13\linewidth}
      \includegraphics[width=\linewidth]{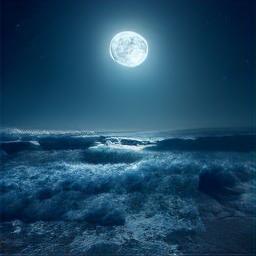}
    \end{subfigure}
    \hspace{0.005\linewidth}
    \begin{subfigure}{0.13\linewidth}
      \includegraphics[width=\linewidth]{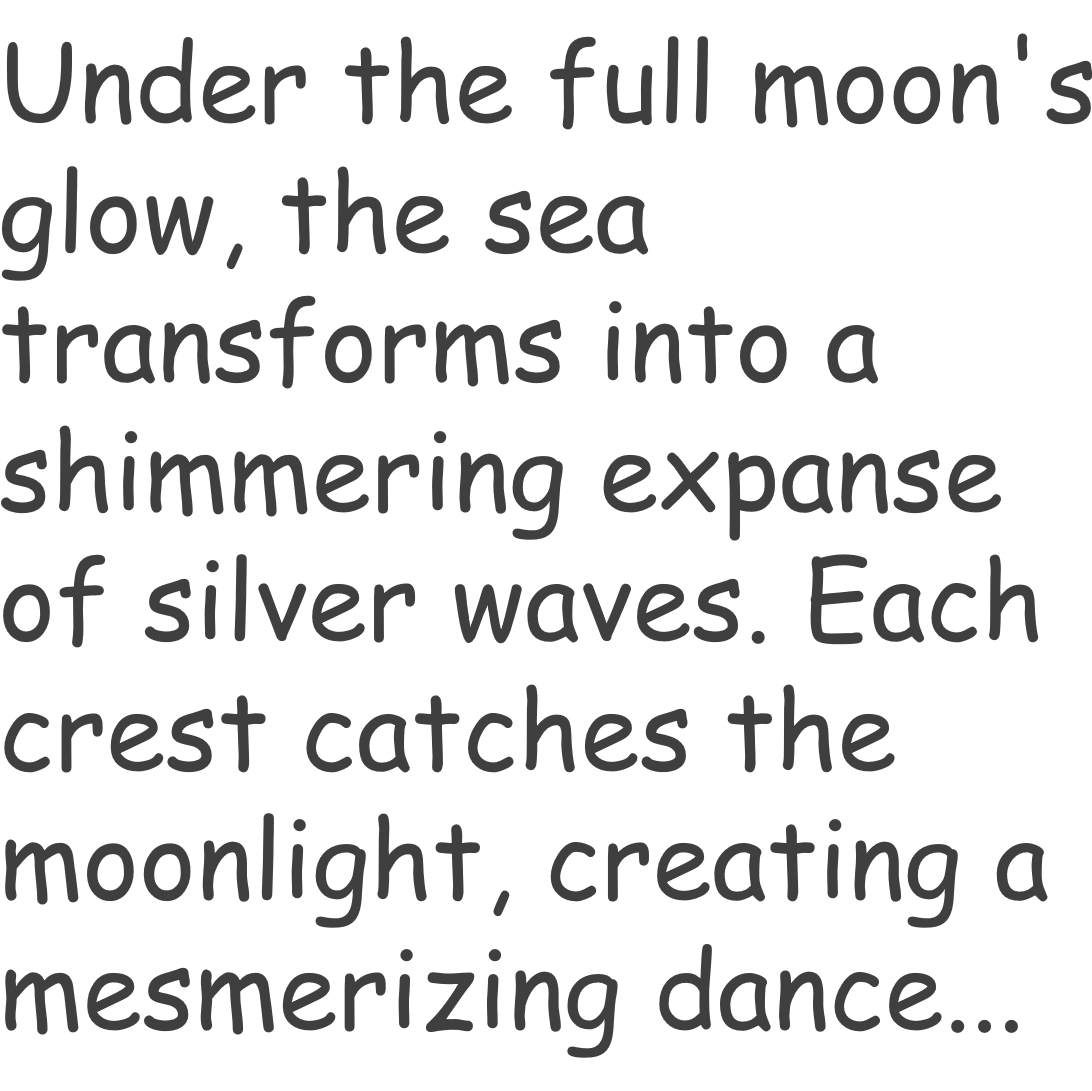}
    \end{subfigure}
    \hspace{0.005\linewidth}
    \begin{subfigure}{0.13\linewidth}
      \includegraphics[width=\linewidth]{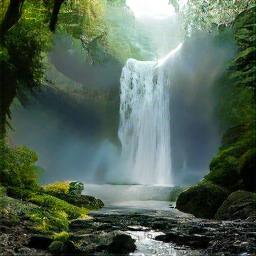}
    \end{subfigure}
    \hspace{0.005\linewidth}
    \begin{subfigure}{0.13\linewidth}
      \includegraphics[width=\linewidth]{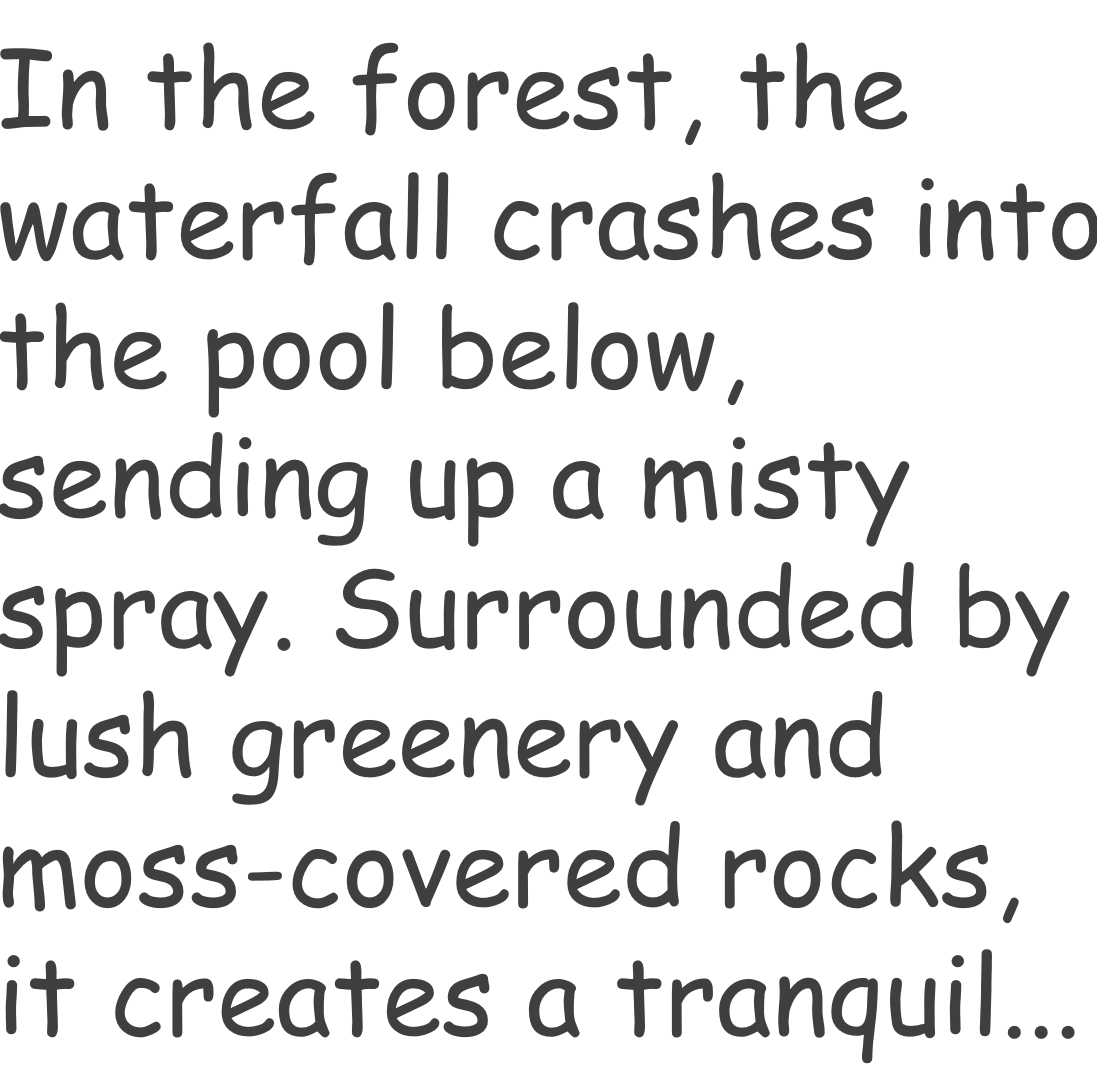}
    \end{subfigure}
  }
  \caption{Examples of model text - to - image (left part, prompts: ``\textit{Please draw me a picture of a man looking at the camera; Could you draw me a picture of a peaceful living room?...} ) and text - to - text and image examples (right part, prompt: ``\textit{I want to know what volcanic lava looks like when it's churning. Could you give me a brief introduction? It would be even better with picture explanations...})}
  \label{fig:show-fig-all}
\end{figure*}

\section{Experiments}
\newcommand{\imgbos}{\texttt{<imgbos>}}
\newcommand{\imgend}{\texttt{<imgend>}}
\newcommand{\imgpad}{\texttt{<imgpad>}}

\subsection{Settings}

\subsubsection{Benchmarks}
To comprehensively evaluate the capabilities of ARMOR, we conduct detailed evaluation experiments from two dimensions: multimodal understanding evaluation and visual generation evaluation.
For evaluating the multimodal understanding ability, we use the VLMEvalkit~\citep{vlmevalkit} platform and evaluate on a series of standardized benchmarks, including MMMU~\citep{mmmu}, MME-P~\citep{mme}, MME~\citep{mme}, MMvet~\citep{mmvet}, MMB~\citep{mmbench}, SEEDBench-img~\citep{seedbench}, HallusionBench~\citep{Hallusionbench}, POPE~\citep{pope} and LLaVABench~\citep{liu2023llava}. These datasets cover a wide range of multimodal understanding tasks, providing a comprehensive evaluation of the model's understanding ability across different contexts.

For evaluating the visual generation capability (\textit{e.g.}, text-to-image generation), we conduct extensive experiments via the GenEval~\citep{geneval} platform. The GenEval platform provides a standardized evaluation framework for text-to-image generation tasks, effectively measuring the quality and accuracy of text-to-image generation. In addition, we also evaluated the model's generation ability based on the MS-COCO ~\citep{mscoco} dataset. We randomly selected 30k questions from it and calculated the FID~\citep{FID} score of the images generated by the model.

\subsubsection{Baselines}

\begin{enumerate}[label=\arabic*., leftmargin=*]
    \item \textit{Understanding models}: These models mainly focus on the understanding ability of multimodal data and can process information from different modalities (such as text and image). Representative models include QwenVL~\citep{Qwen-VL,Qwen2-VL,Qwen2.5-VL}, InternVL~\citep{internvl2.5,chen2024internvl}, InstructBLIP~\citep{blip2,instructblip,li2022blip} and the LLava~\citep{liu2023llava,liu2023improvedllava} series, etc. 
    \item \textit{Generation models}: These models have strong generation ability and can generate high-quality images, including DALLE~\citep{dalle1,dalle2,dalle3}, SDv1.5~\citep{sd3,sdxl}, LlamaGen~\citep{llamagen}, etc. 
    \item \textit{Unified understanding and generation models}: These models not only have multimodal understanding ability but also have good generation ability. Such as Chameleon~\citep{chameleon}, Show-o~\citep{showo},VILA-U~\citep{vilau}, etc.
\end{enumerate}

\subsection{Quantitative Evaluation}

\begin{table*}[!htbp]
    \centering
    \resizebox{\linewidth}{!}{
        \begin{tabular}{l | c c c c c c c c c c c c}
            \toprule
            Method & Params & MMMU (val)~\citep{mmmu} & MME-P~\citep{mme} & MME~\citep{mme} & MMvet~\citep{mmvet} & MMB~\citep{mmbench} & SEEDBench-img~\citep{seedbench} & HallusionBench~\citep{Hallusionbench} & POPE~\citep{pope} & LLaVABench~\citep{liu2023llava} & \\
            \midrule
            \multicolumn{12}{l}{\textit{Understanding modals}} \\
            \midrule
            Qwen2.5-VL & 7B & 56.2 & 1685.2 & 2299.2 & 66.6 & 83.5 & 71.0 & 56.3 & 86.1 & 80.6 \\
            InternVL 2.5 & 8B & 53.5 & 1688.2 & 2338.9 & 59.6 & 82.0 & 77.0 & 49.0 & 88.9 & 80.3 \\
            Qwen2-VL & 7B & 53.7 & 1639.2 & 2276.3 & 61.8 & 82.8 & 76.0 & 50.4 & 88.4 & 70.1 \\
            LLaVA-Next-Vicuna & 13B & 37.3 & 1448.4 & 1745.6 & 44.9 & 70.0 & 71.4 & 31.8 & 87.8 & 73.9 \\
            LLaVA-ov & 7B & 47.9 & 1577.8 & 1993.6 & 51.9 & 83.2 & 76.7 & 31.6 & 88.4 & 81.0 \\
            Llama-3-VILA1.5 & 8B & 37.4 & 1438.8 & 1698.5 & 41.9 & 62.1 & 65.0 & 35.3 & 83.3 & 71.7 \\
            DeepSeek-VL2 & 16B & 54.0 & 1632.7 & 2230.2 & 60.0 & 84.1 & 77.0 & 45.3 & - & 89.7 \\
            LLaVA-v1.5 & 7B & 35.7 & 1506.2 & 1808.4 & 32.9 & 66.5 & 65.8 & 27.6 & 86.1 & 61.8 \\
            InstructBLIP & 7B & 30.6 & 1137.1 & 1391.4 & 33.1 & 33.9 & 44.5 & 31.2 & 86.1 & 59.8 \\
            Qwen-VL-Chat & 7B & 37.0 & 1467.8 & 1860.0 & 47.3 & 61.8 & 64.8 & 36.8 & 74.9 & 67.7 \\
            Emu3\_Chat & 8B & 33.9 & 1334.1 & 1610.5 & 29.1 & 63.8 & 69.2 & 31.7 & 83.3 & 49.2 \\
            \midrule
            \multicolumn{12}{l}{\textit{Uni modals \textbf{without} interleaved text - image output}} \\
            \midrule
            Show-o-256 & 1.3B & 25.1 & 948.4 & - & - & - & - & - & 73.8 & - \\
            SEED-X & 17B & 35.6 & 1435.7 & - & - & - & - & - & 84.2 & - \\
            VILA-U-384 & 7B & - & 1401.8 & - & 33.5 & - & 59.0 & - & 85.8 & - \\
            LWM & 7B & - & - & - & 9.6 & - & - & - & 75.2 & - \\
            TokenFlow-B & 13B & 34.2 & 1353.6 & 1660.4 & 22.4 & - & 60.4 & - & 84.0 & - \\
            TokenFlow-L & 13B & 34.4 & 1365.4 & 1622.9 & 27.7 & - & 62.6 & - & 85.0 & - \\
            TokenFlow-XL-Vicuna & 13B & 38.7 & 1545.9 & 1840.9 & 40.7 & - & 68.7 & - & 86.8 & - \\
            TokenFlow-XL-Qwen & 14B & 43.2 & 1551.1 & 1922.2 & 48.2 & - & 72.6 & - & 87.8 & - \\
            SynerGen-VL & 2.4B & 34.2 & 1381.0 & 1837.0 & 34.5 & 53.7 & 62.0 & - & 85.3 & - \\
            Janus-Pro & 7B & 41.6 & 1516.7 & 1791.7 & 45.1 & 62.6 & 70.1 & 39.5 & 78.9 & 74.4 \\
            \midrule
            \multicolumn{12}{l}{\textit{Uni modals \textbf{with} interleaved text - image output}} \\
            \midrule
            chameleon & 7B & 22.4 & 153.1 & 202.7 & 8.3 & 15.4 & 30.5 & 17.1 & 19.4 & 26.6 \\
            VARGPT & 7B+2B & 36.44 & 1488.8 & - & - & 67.6 & 67.9 & - & 84.4 & - \\
            \textbf{ARMOR (InternVL2.5)} & 8B & \textbf{51.5} & \textbf{1635.2} & \textbf{2281.5} & \textbf{56.3} & \textbf{78.5} & \textbf{75.3} & \textbf{47.6} & \textbf{87.9} & \textbf{78.7} \\
            \bottomrule
        \end{tabular}
    }
    \caption{Evaluation on multimodal understanding benchmarks. We include several methods with their results on multiple benchmarks. The results of ARMOR are highlighted in bold.}
    \label{tab:und_results}
\end{table*}

The experimental results on multimodal understanding benchmarks are shown in Table ~\ref{tab:und_results}.
We can observe that:
(1) Our ARMOR consistently outperforms all existing UniMs across all 9 benchmarks, though they have more parameters (\textit{e.g.}, 51.5 vs 35.6 in score for Ours ARMOR-8B and SEED-X-17B, respectively).
(2) Our ARMOR achieves comparable results with MLLMs. Due to the lack of all the fine-tuning data of InternVL2.5, it is extremely difficult to completely retain its original capabilities. However, we have managed to preserve over $\sim$95\% of its understanding capabilities as much as possible.
The observations highlight the limitations of existing UniMs in multimodal understanding, while demonstrating the significant potential of our framework in endowing existing MLLMs with generation capabilities.

In addition, the experimental results on the visual generation benchmark (\textit{i.e.}, GenEval) and FID score are listed in Table ~\ref{tab:geneval_benchmark_new_2}.
The experimental results demonstrate that ARMOR achieves promising visual generation performance while requiring significantly fewer trainable parameters and lower training costs compared to existing models.
This compelling evidence validates the feasibility of constructing UniMs through upgrading existing MLLMs.

\begin{table*}[!htbp]
\centering
\resizebox{0.8\linewidth}{!}{
\begin{tabular}{llccccccc|c|c}
        \toprule
        Type & Method & \#Param & \#Train Images & Train Cost(GPU days) & Image Res & GenEval  $\uparrow$ & FID $\downarrow$\\
        \midrule
        \multicolumn{1}{l|}{Gen.} & LlamaGen & 0.8B & 60M & - & 256 & 0.32 & 8.69\\
        \multicolumn{1}{l|}{} & LDM & 1.4B & 400M & - & 1024 & 0.37 & 12.64 \\
        \multicolumn{1}{l|}{} & Emu3-Gen & 8B & - & - & 512 & 0.54 & 19.3 \\
        \multicolumn{1}{l|}{} & SDXL & 7B & 2000M & - & 1024 & 0.55 & 9.55\\
        \multicolumn{1}{l|}{} & SDv3 (d=24) & 2B & - & - & 1024 & 0.62 & - \\
        \multicolumn{1}{l|}{} & SDv2.1 & 0.9B & - & 8333/A100 & 768 & 0.50 & 26.96\\
        \multicolumn{1}{l|}{} & SDv1.5 & 0.9B & 2000M & 6250/A100 & 512 & 0.43 & 9.62 \\
        \multicolumn{1}{l|}{} & DALL-E2 & 6.5B & 650M & 4166/A100 & 1024 & 0.52 & 10.39 \\
        \multicolumn{1}{l|}{} & PixArt-alpha & 0.6B & 25M & 753/A100 & 1024 & 0.48 & 7.32\\
        \midrule
        \multicolumn{1}{l|}{NoILO.} & VILA-U & 7B & 15M & - & 384 & 0.42 & 7.69 \\
        \multicolumn{1}{l|}{} & Show-o & 1.3B & 36M & - & 512 & 0.53 & 9.24\\
        \multicolumn{1}{l|}{} & D-DiT & 2B & 400M & - & 512 & 0.65 & - \\
        \multicolumn{1}{l|}{} & TokenFlow-XL & 14B & 60M & - & 384 & 0.55 & -\\
        \multicolumn{1}{l|}{} & SynerGen-VL & 2.4B & 667M & - & 512 & 0.61 & 7.65\\
        \multicolumn{1}{l|}{} & Janus-Pro-7B & 7B & 72M & 3584/A100 & 384 & 0.80 & -\\
        \multicolumn{1}{l|}{} & Janus-Pro-1B & 1.5B & 72M & 1568/A100 & 384 & 0.73 & -\\
        \multicolumn{1}{l|}{} & SEED-X & 17B & 158M & $\sim$ 960/A100 & - & 0.49 & 14.99\\
        \midrule
        \multicolumn{1}{l|}{ILO.} & Chameleon & 7B & 1.4B & 35687/A100 & 512 & 0.39 & -\\
        \multicolumn{1}{l|}{} & ARMOR & 8B & 5M & $\sim$ 500/A100 & 256 & 0.51 & 9.07\\
        \bottomrule
\end{tabular}
}
\caption{GenEval and FID score. Gen. denotes ``generation'' and NoILO. denotes ``UniMs \textbf{without} interleaved text-image output''. ILO. denotes ``UniMs \textbf{with} interleaved text-image output''. The ``$\sim$''  represents the conversion of computing power. H100 is approximately 2.5 times that of A100, and H800 is approximately 2 times that of A100.}
\label{tab:geneval_benchmark_new_2}
\end{table*}

\subsection{Qualitative Evaluation}

First, we verified the necessity of the second training stage through qualitative evaluation. The experimental results are shown in Figure \ref{fig:show-fig}. It can be observed that for the two different prompts in lines 1 and 2, as the number of training epochs in the second training stage increases, the quality of the generated images improves accordingly. This indicates that the second training stage (i.e., the ``how to generate'' stage) significantly enhances the image quality generated by the multimodal large language model. Second, the output of the model trained in the third stage is shown in Figure \ref{fig:show-fig-all}. The model can make correct responses in different situations (for example: "Please help me draw an apple", such a simple question that only requires a single image response; "Have you seen what the frozen lake looks like in the early morning? Please draw a picture to show it as well", such a question that requires both text and image output). Finally, the analysis results demonstrate the effectiveness of our proposed WoHG training method. In addition, we enabled MLLMs to acquire image generation capabilities with almost no loss of their original capabilities, which also proves the feasibility of the ARMOR architecture and the potential of MLLMs for unified understanding and generation.

\begin{figure}[t]
  \centering
   \includegraphics[width=0.95\linewidth]{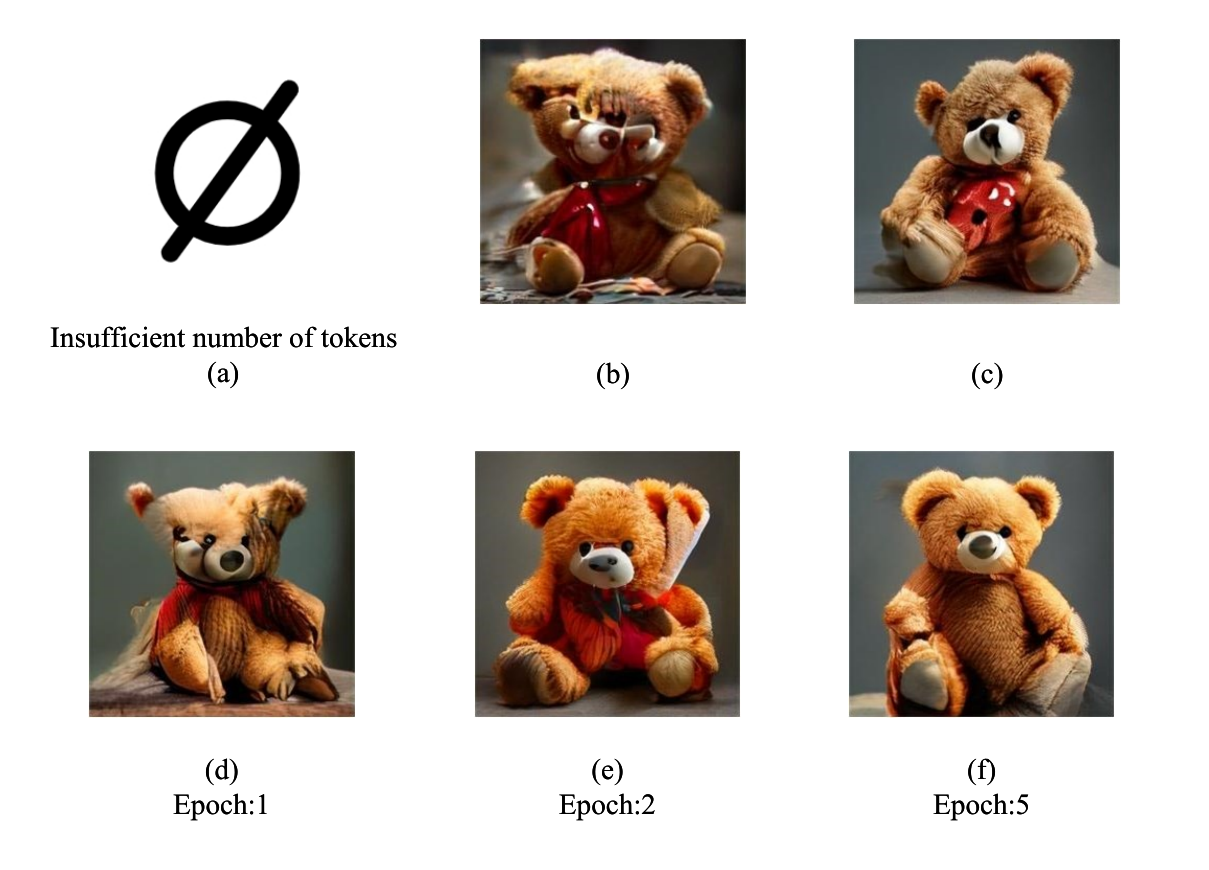}
   \caption{Generated images with the prompt ``\textit{Can you help me draw a picture of a teddy bear doll?}". Images (a), (b) and (c) are from a model with one output layer, and images (d), (e) and (f) are from a model with two output layers.}
   \label{fig:loss-fig}
   \vspace{-4mm}
\end{figure}

\section{Ablation Studies}

\subsection{Multimodal Understanding Experiment}
\vspace{-1mm}

To verify the effectiveness of the training method in preserving the original capabilities, we designed an ablation experiment for the model's multimodal understanding ability. We divided the trainable modules of the model (excluding the visual tower and the connector) into 6 parts. In the experiment, we trained different combinations of modules. The ratio of the datasets used for training was: t2t: 100k, ti2t: 300k, t2i: 100k, t2ti: 100k. The hyperparameters used in the training were consistent with those in ~\ref {sec:training}. We used the three benchmark scores of MMMU, MME, and MMB as indicators to measure the model's capabilities. The experimental results are shown in Table ~\ref{tab:understanding}. Due to the lack of all the fine-tuning data of the base model, the participation of the model's decoder layer and the original embedding module in training would cause a significant loss of capabilities, while the other four modules have a relatively small impact on the model's capabilities, proving the effectiveness of the training method.

\begin{table*}[!htbp]
    \vspace{-4mm}
    \caption{Experimental results of model capabilities under different training methods. Swith represents the model's ability to correctly select the response mode according to the question type.} 
    \label{tab:understanding}
    \centering
    \resizebox{\linewidth}{!}{
    \begin{tabular}{c|cccccc|c|ccc}
        \toprule
        \multirow{2}{*}{Exp.} & \multicolumn{6}{c}{Training parameters} & \multicolumn{4}{|c}{Result} \\
        \cmidrule{2-11}
         & \makecell{InternLM embedding} & \makecell{Added embedding} & \makecell{Decoder layer} & \makecell{Text output} & Adapter & \makecell{Visual output} & Switch & MMMU & MME & MMB \\
        \midrule
        \multicolumn{1}{l|}{1}  & \checkmark & \checkmark & \checkmark & \checkmark & \checkmark & \checkmark & \checkmark & 36.8 & 1532.3 & 57.5\\
        \multicolumn{1}{l|}{2}  &            & \checkmark & \checkmark & \checkmark & \checkmark & \checkmark & \checkmark & 38.9 & 1617.4 & 65.3 \\
        \multicolumn{1}{l|}{3}  & \checkmark & \checkmark &            & \checkmark & \checkmark & \checkmark & \checkmark & 35.6 & 1497.7 & 65.1 \\
        \rowcolor[HTML]{E0F2F1}
        \multicolumn{1}{l|}{4}  &            & \checkmark &            & \checkmark & \checkmark & \checkmark & \checkmark & 50.9 & 2273.7 & 78.8 \\
        \rowcolor[HTML]{E0F2F1}
        \multicolumn{1}{l|}{5}  &            & \checkmark &            & \checkmark &            &            & \checkmark & 51.5 & 2281.1 & 78.5 \\
        \multicolumn{1}{l|}{6}  &            &            &            & \checkmark &            &            &            & 52.2 & 2220.5 & 77.9 \\
        \multicolumn{1}{l|}{7}  &            & \checkmark &            &            &            &            &            & 53.1 & 2339.4 & 81.7 \\
        \bottomrule
    \end{tabular}
    }
\vspace{-1mm}
\end{table*}

\subsection{Experiment on Forward-switching Mechanism}

To verify the promoting effect of the forward-switching mechanism proposed in this paper on image generation training. We used a dataset of 0.5 million samples (t2i: 300k; t2ti: 200k) to train the model with the forward-switching mechanism and the model with the initial structure for 5 epochs. The training hyperparameters were set to be the same as those in Section ~\ref{sec:training}. The performance of image generation during the training process is shown in Figure \ref{fig:loss-fig}. Judging from the experimental results, due to the absence of the long-tail distribution problem, under the same training intensity, the model with the forward-switching mechanism performs better. It is worth noting that after one epoch of learning, the forward-switching mechanism model can correctly output the number of tokens and thus generate images normally, while the initial structure model may have a phenomenon where the output tokens contain a mixture of images and text, resulting in generation failure. This indicates that the forward-switching mechanism can effectively improve the model's learning rate, demonstrating the positive effect of the proposed method.

\subsection{Experiment on Model Scale}

To verify whether the model's generation ability conforms to the scaling law, we designed the following experiments: Set the number of adapter layers of the model to 2 layers (2a) and 4 layers (4a) respectively. The training data is 0.3M (t2i, which is the same as the dataset used in the third stage of WoHG). Continuously test the GenEval index and FID (COCO - 30k) scores of the models trained in step1 of the WoHG method (S1) and those trained in both step1 and step2 (S2) during the training process. The training hyperparameters are kept consistent with those in step3 of section ~\ref{sec:training}. The experimental results are recorded in the table. Analyzing the experimental results, we can see that the smaller-scale model shows a faster learning speed, but the larger - scale model shows higher learning quality. On the other hand, the geneval scores also prove this fact. In addition, by comparing the models with the two training strategies, we can demonstrate the effect of the second - stage training on improving the model's ability. Finally, the experimental results confirm the application potential of the method proposed in this paper when expanding the model scale.

\begin{table}[htbp]
    \centering
    \resizebox{\linewidth}{!}{
    \begin{tabular}{c|c|c|c|cccc}
        \toprule
        \multirow{2}{*}{Result} & \multirow{2}{*}{Training} & \multirow{2}{*}{Model}  & \multirow{2}{*}{Param} & \multicolumn{4}{c}{Training Progress} \\
        \cmidrule{5-8}
         &  &  &  & 10\% & 30\% & 50\% & 100\% \\
        \midrule
        \multicolumn{1}{l|}{FID$\downarrow$}  & S1 & 2a & 0.5B & 57.67 & 37.11 & 24.26 & 19.35 \\
        \multicolumn{1}{l|}{}  & S1 & 4a & 0.9B & 65.52 & 42.77 & 26.67 & 16.33 \\
        \multicolumn{1}{l|}{}& \cellcolor[HTML]{E0F2F1}S2 & \cellcolor[HTML]{E0F2F1}2a & \cellcolor[HTML]{E0F2F1}0.5B & \cellcolor[HTML]{E0F2F1}25.01 &\cellcolor[HTML]{E0F2F1}17.32 &\cellcolor[HTML]{E0F2F1}12.19 & \cellcolor[HTML]{E0F2F1}9.59 \\
        \multicolumn{1}{l|}{}& \cellcolor[HTML]{E0F2F1}S2 & \cellcolor[HTML]{E0F2F1}4a & \cellcolor[HTML]{E0F2F1}0.9B & \cellcolor[HTML]{E0F2F1}19.31 &\cellcolor[HTML]{E0F2F1}14.2 &\cellcolor[HTML]{E0F2F1}11.08 & \cellcolor[HTML]{E0F2F1}9.07 \\
        \midrule
        \multicolumn{1}{l|}{GenEval$\uparrow$} & S1 & 2a & 0.5B & 0.15 & 0.21 & 0.26 & 0.28 \\
        \multicolumn{1}{l|}{} & S1 & 4a & 0.9B & 0.13 & 0.19 & 0.26 & 0.31 \\
        \multicolumn{1}{l|}{}&\cellcolor[HTML]{E0F2F1}S2 & \cellcolor[HTML]{E0F2F1}2a & \cellcolor[HTML]{E0F2F1}0.5B &\cellcolor[HTML]{E0F2F1}0.33 &\cellcolor[HTML]{E0F2F1}0.39 &\cellcolor[HTML]{E0F2F1}0.43 &\cellcolor[HTML]{E0F2F1}0.47 \\
        \multicolumn{1}{l|}{}&\cellcolor[HTML]{E0F2F1}S2 & \cellcolor[HTML]{E0F2F1}4a & \cellcolor[HTML]{E0F2F1}0.9B &\cellcolor[HTML]{E0F2F1}0.35 &\cellcolor[HTML]{E0F2F1}0.42 &\cellcolor[HTML]{E0F2F1}0.46 &\cellcolor[HTML]{E0F2F1}0.51 \\
        \bottomrule
    \end{tabular}
    }
    \label{tab:answer-tab}
\end{table}
\vspace{-5mm}

%% file: sec/5_conclusion.tex
\section{Conclusion}

In this paper, we have presented the ARMOR framework to build UniMs by upgrading existing MLLMs.
By introducing an asymmetric encoder-decoder architecture, the framework enables MLLMs to output natural interleaved text-image.
We have collected a high-quality interleaved dataset and developed a three-stage training algorithm named WoHG to fine-tune existing MLLMs.
The algorithm enables MLLMs to achieve unified capabilities in both understanding and generation tasks.
Experimental results on 9 benchmarks demonstrate that our framework can effectively empower existing MLLMs with generation capabilities while preserving their understanding capabilities.

%% file: sec/X_suppl.tex
\clearpage
\setcounter{page}{1}
\maketitlesupplementary

\section{Appendix}
\label{sec:rationale}

\subsection{Dataset Examples}

\noindent \textbf{1. t2t (Text-to-Text) Example:}

\vspace{6pt}

\noindent \textbf{From:} human

\noindent \textbf{Value:} What kind of scenery can be found beside the Nile River? 

\vspace{6pt}

\noindent \textbf{From:} gpt

\noindent \textbf{Value:} The Nile River offers desert landscapes, lush farmlands, palm trees, small villages, ancient monuments, wildlife, and the fertile Nile Delta, creating a mix of natural beauty and historical significance.

\vspace{12pt}

\noindent \textbf{2. t2ti (Text-to-Text-Image) Example:}

\vspace{6pt}

\noindent \textbf{From:} human

\noindent \textbf{Value:} What kind of scenery can be found beside the Nile River? Can you draw an image to depict it for me?

\vspace{6pt}

\noindent \textbf{From:} gpt

\noindent \textbf{Value:} Alongside the Nile River, you can find lush, verdant banks filled with vibrant greenery and towering papyrus plants. The scene reflects a serene environment with occasional traditional felucca sailboats gliding across the water.

\vspace{6pt}

\begin{center}
    \includegraphics[width=0.2\textwidth]{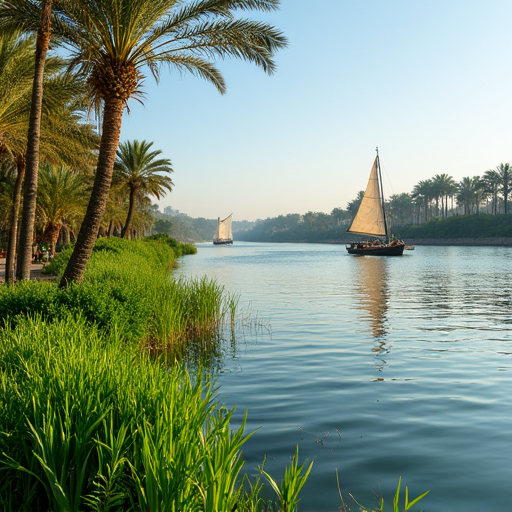}
    \label{fig:t2timage}
\end{center}

\vspace{12pt}

\noindent \textbf{3. t2i (Text-to-Image) Example:}

\vspace{6pt}

\noindent \textbf{From:} human

\noindent \textbf{Value:} Please draw an image of a sunset over the sea for me.

\vspace{6pt}

\noindent \textbf{From:} gpt

\noindent \textbf{Value:}

\vspace{6pt}

\begin{center}
    \includegraphics[width=0.2\textwidth]{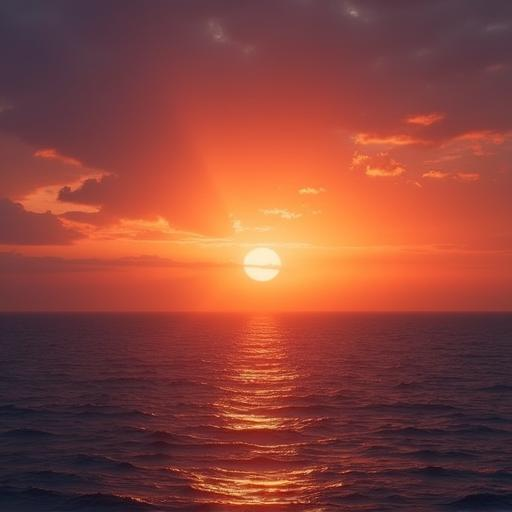}
    \label{fig:t2iimage}
\end{center}

\vspace{12pt}

\noindent \textbf{4. ti2t (Text-Image-to-Text) Example:}

\vspace{6pt}

\noindent \textbf{From:} human

\noindent \textbf{Value:} \textless\text{image}\textgreater What is this photo about?

\vspace{6pt}

\begin{center}
    \includegraphics[width=0.2\textwidth]{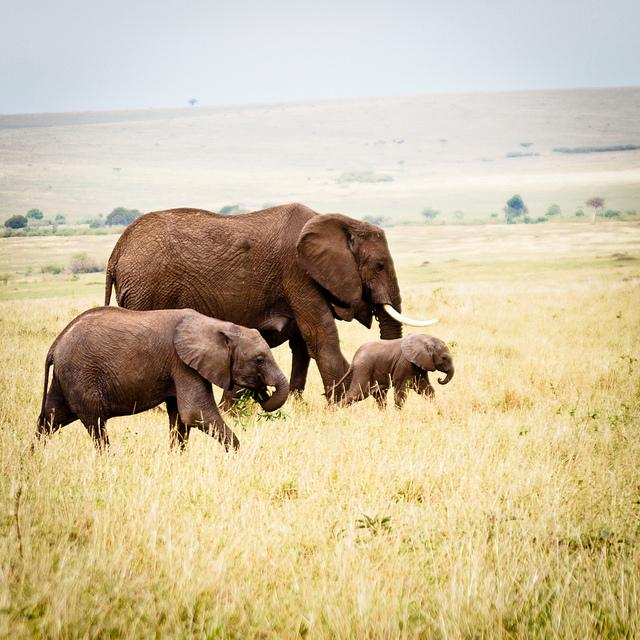}
    \label{fig:ti2timage}
\end{center}

\vspace{6pt}

\noindent \textbf{From:} gpt

\noindent \textbf{Value:} In the heart of a vast grassland, a family of three elephants is on a journey. The adult elephant, with its majestic brown skin, leads the way in the center, its size significantly larger than the two smaller ones. On either side of this gentle giant, slightly behind, are two smaller elephants, their youthful energy evident in their stride. The elephants are making their way through a field of tall grass, their brown bodies contrasting beautifully with the greenery around them. The grassland extends far into the distance, dotted here and there with a few trees standing tall and solitary. Above them, the sky is a canvas of overcast clouds, casting a soft light over the scene. The relative positions of the elephants suggest a protective formation, with the adult elephant guiding its young ones through the grassland. This image captures a moment of tranquility and familial bond in the wild.

\subsection{Data Ratio in the Training Phase}

\textbf{In the first stage of training}, the experiment was set to train the newly added embedding layer and the output layer of internvl. Since training the output layer of the internvl model is bound to affect the original capabilities of the model, how to reduce the degree of this negative impact has become one of the key issues we need to solve. We tested and found that using the training scheme of t2t (100K), ti2t (100K), t2i (100K) and t2ti (100K) still reduces the original capabilities of the model, resulting in a worse instruction - following ability when the model faces some questions. Therefore, we carefully selected 300K data from the open - source dataset (mainly LLaVA - NEXT) for model training, so that the degree of decline in the original capabilities of the model after the first - stage training becomes very small. 

\textbf{In the second stage of training}, the experiment was set to only train the newly added parts. Therefore, we used a large amount of t2i and t2ti data to train the generation ability of the model without having to worry about the impact of training new capabilities on the original capabilities. 

\textbf{In the third stage of training}, in the experiment, the newly added embedding layer, the newly added adapter, the output layer of internvl, and the final visual output layer were set to be trained. Initially, the data ratio we used was t2t(50K), ti2t (50K), t2i (50K) and t2ti (50K). Due to the low - quality t2ti data further reducing the original ability level of the model, and the small scale of the image fine - tuning data volume, the final geneval ability score of the model was only 0.37. After we expanded t2i to (300k), we found that its geneval score reached 0.47. Furthermore, when we increased the newly added transformer layers of the model to 4 layers, we obtained a generation result of 0.51, and we believe there is still room for improvement. In addition, when we also expanded the ti2t data to 300k, compared with the initial training scheme, the MMMU score increased from 49.8 to 51.5, and there were also improvements to varying degrees in other dimensions of the benchmark.  

\subsection{Work motivation and details}
The purpose of ARMOR is to transform a pre-trained MLLM into a model with unified understanding and generation capabilities. Currently, unified understanding and generation models often require a large amount of data to train the model's multimodal understanding abilities. Some models attempt to achieve this function by using an external attachment approach, such as Metamorph and SEED-X. These two models are also implemented based on pre-trained models. However, during the training process, they rely on the original capabilities of the model for training. Although this can reduce the amount of training, it undoubtedly still damages the original capabilities of the model. ARMOR completely "stands on the shoulders of giants" to expand its capabilities. During the training process, we found that it is extremely difficult to retain the original capabilities of the model. Even when we only train a small part of the model's parameters, it can cause catastrophic damage to the model's capabilities. To this end, we have made a lot of efforts to retain the original capabilities of the model. In the first stage of training in this work, our initial goal was simple, which was to make the model learn to correctly generate the switch of the image modality (\texttt{<imgbos>}) according to the question. However, although the output of the model is normal after training, its performance on the benchmark almost collapses. At the same time, the inability to obtain all the fine-tuning data of the model is a challenge we face. Therefore, we have screened from open-source datasets and used the distillation method to obtain a batch of high-quality image-text understanding datasets. This can retain the original capabilities of the model to the greatest extent, but it will still cause some loss of capabilities. In future research, we will continue to explore more effective methods to make up for the current shortcomings.  

\subsection{Mixed-modality Output Test}

In this subsection, we list some examples of ARMOR's interleaved input and output, as well as comparisons with other models.

\tcbset{
    questionbox/.style={
        colframe=blue!70!black,
        colback=blue!10,
        fonttitle=\bfseries,
        title=Question,
        boxrule=1pt,
        arc=4pt,
        outer arc=4pt,
    },
    answerbox/.style={
        colframe=green!70!black,
        colback=green!10,
        fonttitle=\bfseries,
        boxrule=1pt,
        arc=4pt,
        outer arc=4pt,
    }
}

\subsubsection{Text-Image-to-Text-Image Test}

\begin{tcolorbox}[questionbox]
    Does the picture show the scene of athletes exercising? Can you help me draw one similar to this picture?
    \begin{center}
        \includegraphics[width=0.6\textwidth]{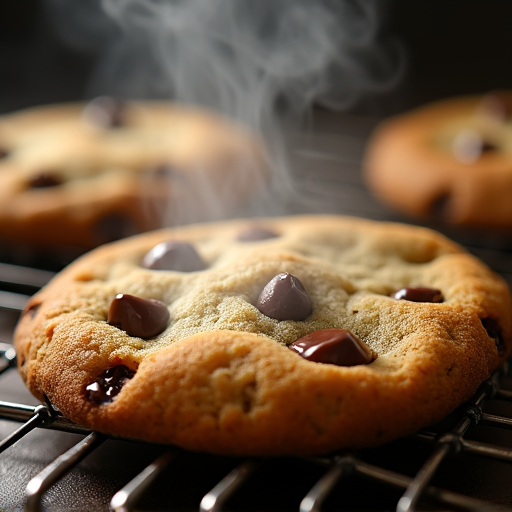}
        \label{fig:X_hallucination_biscuit}
    \end{center}
\end{tcolorbox}

\begin{tcolorbox}[answerbox, title=ARMOR]
    No, the picture does not show athletes in action. It depicts freshly baked chocolate chip cookies cooling on a wire rack. 
    \begin{center}
        \includegraphics[width=0.6\textwidth]{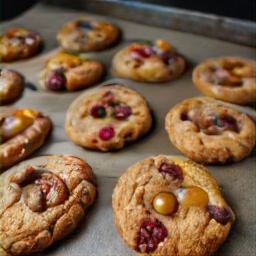}
        \label{fig:X_hallucination_biscuit_ARMOR}
    \end{center}
\end{tcolorbox}

\begin{tcolorbox}[answerbox, title=Janus-Pro]
    The picture you provided shows freshly baked chocolate chip cookies cooling on a wire rack. There are no athletes or exercise scenes in this image. I can't help you draw one similar to this picture as I am an AI language assistant and not capable of visual creation. If you have any other questions or need assistance with a different topic, feel free to ask!
\end{tcolorbox}

\begin{tcolorbox}[answerbox, title=VILA-U]
    No, the picture does not show the scene of athletes exercising. It is a close-up image of a freshly baked cookie on a cooling rack. The cookie is round in shape and has chocolate chips on top.
\end{tcolorbox}

\begin{tcolorbox}[answerbox, title=Show-o]
    {\color{red}Yes, the image shows a scene of athletes exercising, with a person holding a pair of scissors} and a doughnut with chocolate frosting and sprinkles on a cooling rack. The doughnut is placed in the foreground, while the athlete is in the background. The athlete is holding the scissors, possibly preparing to cut the doughnut. The scene is set against a black background, which emphasizes the focus on the athlete and the doughnut.
\end{tcolorbox}

\begin{tcolorbox}[answerbox, title=Anole]
    {\color{red}Gentle climate, nature-friendly outdoor activities, or simply cozy home comforts — the image triggers a sense of wellbeing in viewers. A partially visible sports coach calmly leads his group as they work through a warm-up exercise on what looks like a quaint sports field.} Freshly baked donuts lay ready on the grill, tempting the trained bods to break their routines with a sweet treat. The scene is both focused and relaxed, with each athlete having space to move and enjoy the time spent together.
    Here's an image based on the provided picture:
    \begin{center}
        \includegraphics[width=0.6\textwidth]{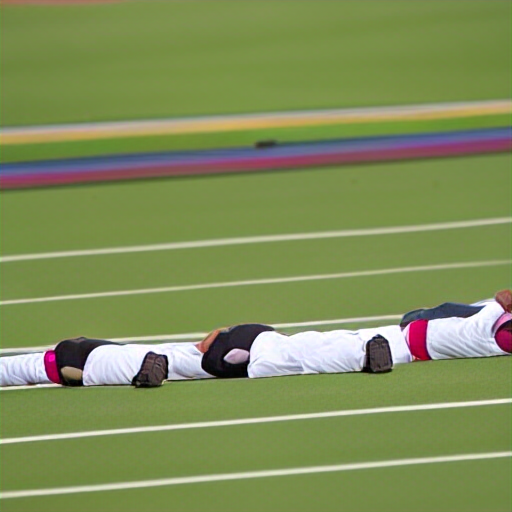}
        \label{fig:X_hallucination_biscuit_ARMOR_1}
    \end{center}
\end{tcolorbox}

\begin{tcolorbox}[answerbox, title=VARGPT]
    No, the picture does not show athletes exercising. It shows a close-up of a chocolate chip cookie on a cooling rack. 
\end{tcolorbox}

This is a evaluation integrating understanding and generation requirements. Through the evaluation, Janus-Pro can recognize that the image doesn’t show athletes but a cookie, and responds to the subsequent image-generation request by clearly stating it’s a language model unable to perform visual generation. 
VILA-U can also identify that it’s not an athlete scene, yet makes no response to the image-generation demand.
However, Show-o fails to accurately understand the image content. It even describes a false image caption in a seemingly plausible way and doesn’t respond to the image-generation request at all. 
Compared with Janus-Pro, VILA-U and Show-o, ARMOR can accurately understand the image content, confirm it’s not about athletes, and precisely recognize the image-generation requirement, generating an image that meets the needs.Although Anloe stated in the latter part of the answer that these were freshly baked cookies, in the first part of the answer, it was still describing scenes of athletes, and the image drawn was also related to athletes, indicating its deficiency in multimodal understanding ability. VARGPT can correctly identify the confusing question and provide a correct answer, but the probability of it generating a picture is too low. After 20 rounds of testing, it still didn't generate any images, indicating that it automatically ignored the drawing question in the title: Can you help me draw one similar to this picture?

\subsubsection{Text-to-Text-Image Test}

\begin{tcolorbox}[questionbox]
    What's the Northern Hemisphere like in June? Can you draw an image about it for me? 
\end{tcolorbox}

\begin{tcolorbox}[answerbox, title=ARMOR]
    In the Northern Hemisphere during June, it's typically summer with warmer temperatures and longer daylight hours. Regions closer to the poles experience the midnight sun, where the sun remains visible at midnight due to the region's tilt towards the sun. Plants are lush, and wildlife is active 
    \begin{center}
        \includegraphics[width=0.6\textwidth]{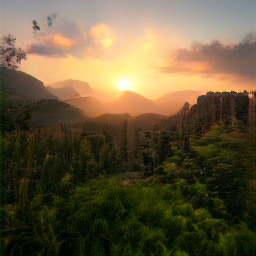}
        \label{fig:X_northern_ARMOR}
    \end{center}
\end{tcolorbox}

\begin{tcolorbox}[answerbox, title=Janus-Pro]
    \begin{center}
        \includegraphics[width=0.6\textwidth]{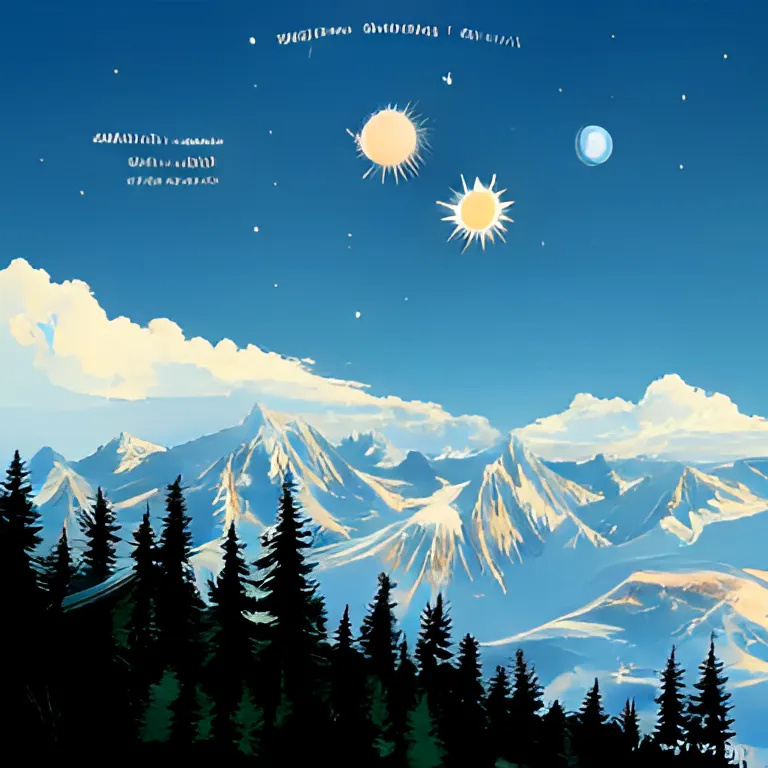}
        \label{fig:X_northern_Janus-Pro}
    \end{center}
\end{tcolorbox}

\begin{tcolorbox}[answerbox, title=VILA-U]
    \begin{center}
        \includegraphics[width=0.6\textwidth]{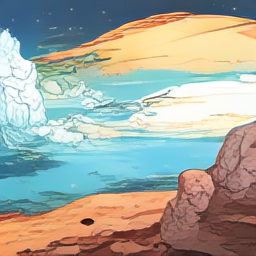}
        \label{fig:X_northern_VILA-U}
    \end{center}
\end{tcolorbox}

\begin{tcolorbox}[answerbox, title=Show-o, arc=3pt, outer arc=3pt]
    \begin{center}
        \includegraphics[width=0.6\textwidth]{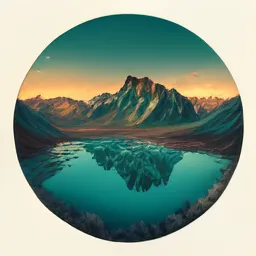}
        \label{fig:X_northern_Show-o}
    \end{center}
\end{tcolorbox}

\begin{tcolorbox}[answerbox, title=Anole]
    June is a beautiful time to visit the Northern Hemisphere. As the spring weather continues to improve, the region is transformed into a lush, vibrant landscape teeming with life. In the United States, Canada, and Mexico, the sun is shining brightly and temperatures are starting to rise. The days are long and warm, making it the perfect season for outdoor activities such as hiking, camping, and exploring national parks.
    One of the most noticeable changes in the Northern Hemisphere in June is the emergence of new vegetation. As the snow and ice begin to melt, plants and trees start to sprout up, filling the air with their sweet scent and vibrant colors. In the United States, for example, the blooming of wildflowers is a particularly stunning sight, with fields of purple lupines, yellow black-eyed Susans, and pink wild roses adding a pop of color to the landscape.
    Another significant change in the Northern Hemisphere in June is the return of wildlife. Many animals, such as birds, deer, and bears, emerge from their winter dens and make their way back to their habitats. In the United States, for example, birdwatchers can enjoy the spectacle of migrating flocks of geese, ducks, and pelicans flying overhead, while deer can be spotted grazing in fields and forests.
    In addition to the return of wildlife, June also marks the beginning of the summer season in the Northern Hemisphere. This means that many popular tourist destinations are now open and ready for visitors. In the United States, for example, popular destinations such as beaches, theme parks, and national parks are all in full swing, providing opportunities for outdoor activities, relaxation, and cultural experiences.
    Overall, June is a wonderful time to visit the Northern Hemisphere. The warm weather, vibrant landscapes, and return of wildlife make it a season of renewal and growth, perfect for exploring the outdoors and experiencing the beauty of the natural world.
    \begin{center}
        \includegraphics[width=0.6\textwidth]{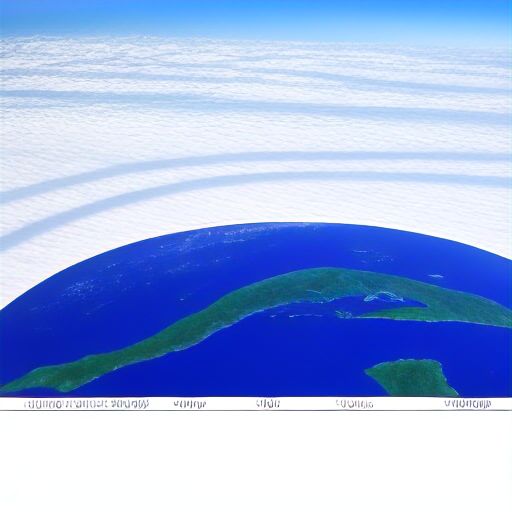}
        \label{fig:X_northern_ARMOR_1}
    \end{center}
\end{tcolorbox}

\begin{tcolorbox}[answerbox, title=VARGPT]
    I created a drawing of the Northern Hemisphere in June. chatGPT.mm
    \begin{center}
        \includegraphics[width=0.6\textwidth]{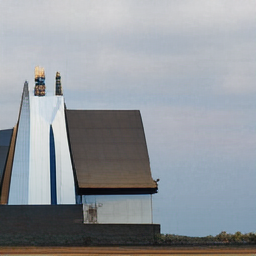}
        \label{fig:X_northern_ARMOR_2}
    \end{center}
\end{tcolorbox}

This evaluation focuses on multimodal generation tasks that require both text descriptions and corresponding images to fully address user requests. Unlike standard image generation, successful responses must provide accurate textual explanations paired with visually appropriate images.

Let's take a look at the output of these models. Janus - Pro, VILA - U, and Show - o can only generate a single image, while ARMOR, VARGPT, and Anole can output responses with alternating text and images. Both Janus - Pro and VILA - U generated snowy mountain scenes. Since June usually represents summer, these depictions of snowy mountains may not meet people's expectations of common summer scenes in June. Additionally, neither of these two images presents real natural landscapes. The image of Janus - Pro adopts a retro illustration style, while the image of VILA - U uses a flat illustration style. Now, look at Show - o. It generated an image of natural scenery, but it misinterpreted the word "hemisphere". The image it drew is circular, with hardly any elements showing summer characteristics. Anole correctly described the scene in the Northern Hemisphere in June, but its output is too long - winded, and the information presented in the image is difficult to distinguish. Although VARGPT demonstrated the ability to output alternating text and images, its text answer has no relation to the question at all, and the image depicts a house, which also doesn't match the question. In contrast, the text output of ARMOR is of moderate length, and it can output an image that highly matches the text content. 

\begin{tcolorbox}[questionbox]
    What's the Southern Hemisphere like in June? Can you draw an image about it for me?
\end{tcolorbox}

\begin{tcolorbox}[answerbox, title=ARMOR]
    In June, the Southern Hemisphere is experiencing winter, with varying conditions across different regions. High latitudes may see continuous daylight or polar night depending on their proximity to the poles. Temperate zones will have cold days typical of winter, with possible snowfall in mountainous areas. Tropical regions may experience rainy seasons with warmer temperatures.
    \begin{center}
        \includegraphics[width=0.6\textwidth]{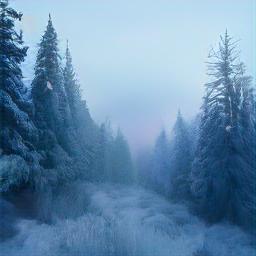}
        \label{fig:X_southern_ARMOR}
    \end{center}
\end{tcolorbox}

\begin{tcolorbox}[answerbox, title=Janus-Pro]
    \begin{center}
        \includegraphics[width=0.6\textwidth]{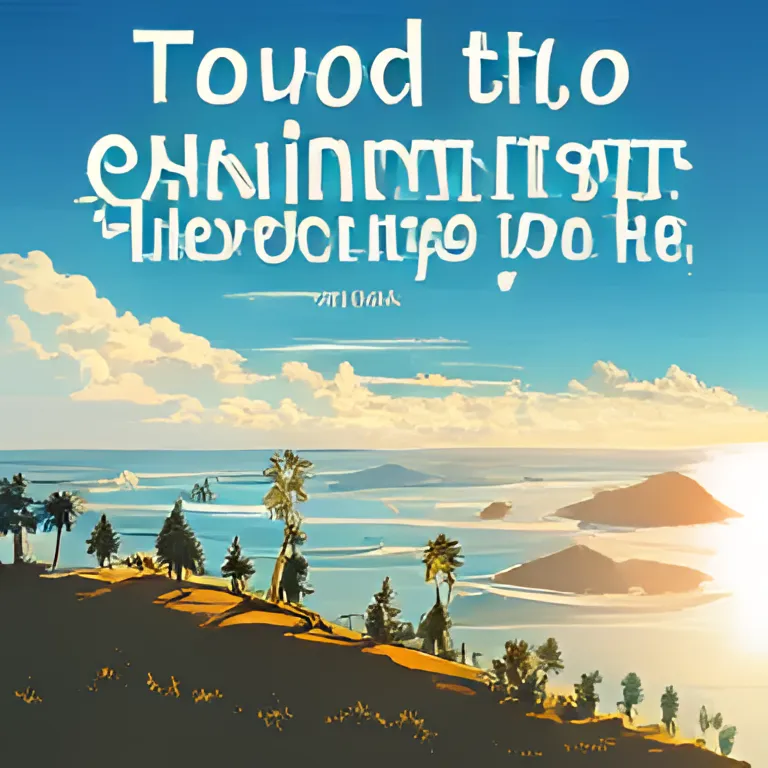}
        \label{fig:X_southern_Janus-Pro}
    \end{center}
\end{tcolorbox}

\begin{tcolorbox}[answerbox, title=VILA-U]
    \begin{center}
        \includegraphics[width=0.6\textwidth]{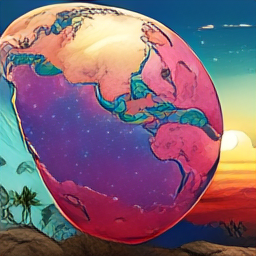}
        \label{fig:X_southern_VILA-U}
    \end{center}
\end{tcolorbox}

\begin{tcolorbox}[answerbox, title=Show-o]
    \begin{center}
        \includegraphics[width=0.6\textwidth]{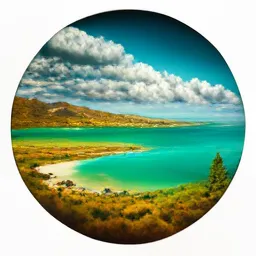}
        \label{fig:X_southern_Show-o}
    \end{center}
\end{tcolorbox}

\begin{tcolorbox}[answerbox, title=Anole]
    {\color{red}In June, the southern hemisphere experiences a transitional season between spring and summer, with mild temperatures and changing weather patterns. The weather can vary widely depending on the specific location and time of year, but generally speaking, June is a pleasant month for exploring the southern hemisphere.}
    Below is a depiction of what the weather might be like in June for each location, using a color key to represent the average temperature for that area:
    * New Zealand (except the far south): Mild and pleasant, with average temperatures in the mid-10s Celsius (50s Fahrenheit).
    * Australia (except for the far north): Warm and sunny, with average temperatures in the mid-20s Celsius (70s Fahrenheit).
    * Southern Africa (excluding the far north): Warm and dry, with average temperatures in the mid-20s Celsius (70s Fahrenheit).
    * South America (except for the far north): Mild and pleasant, with average temperatures in the mid-10s Celsius (50s Fahrenheit).
    Keep in mind that these are general averages and weather patterns can vary significantly from year to year. It's always a good idea to check the current weather forecast before planning your trip.
    \begin{center}
        \includegraphics[width=0.6\textwidth]{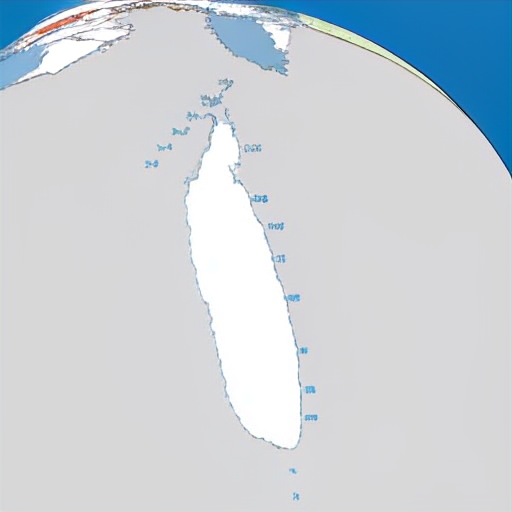}
        \label{fig:X_southern_ARMOR_1}
    \end{center}
\end{tcolorbox}

\begin{tcolorbox}[answerbox, title=VARGPT]
    {\color{red}I'm sorry, but I am not able to create a drawing of the Southern Hemisphere in June. chatGPT. I am a text-based AI model and do not have the capability to create visual images.} However, I can provide you with information about the Southern Hemisphere in June.
    In June, the Southern Hemisphere is in the midst of winter. The southernmost part of the world, including Antarctica, is experiencing the longest and darkest month of the year. The days are short, with the sun barely visible above the horizon, and the nights are long and cold.
    Temperatures in the Southern Hemisphere in June range from freezing in Antarctica to warm in the northern parts of South America and Africa. In general, the weather is cooler and drier in the southern parts of the hemisphere, while the northern parts are warmer and more humid.
    The Southern Hemisphere is also home to several summer festivals and events in June, including the Sydney Festival in Australia, the Stellenbosch Chenin Blanc Festival in South Africa, and the Calla Lily Festival in Chile.
\end{tcolorbox}

This question simply replaces the Northern Hemisphere in the previous example with the Southern Hemisphere. Through testing, neither Janus-Pro, VILA-U, nor Show-o generated the pictures that met the requirements. Although the works generated by Janus-Pro and Show-o are of good quality, they do not match the scene in the Southern Hemisphere in June. VILA-U tried to draw a globe, which not only fails to meet the requirements of the question but also has poor generation quality. In addition, the picture drawn by Show-o is still circular. Anole was able to output content with interleaved images and text, but there was a common-sense error in the first part of its text answer, which still indirectly shows the problem of Anole's relatively weak comprehension ability. And VARGPT showed a phenomenon of refusing to answer before starting the text answer. After our 20 attempts, including modifying the prompt to make it more in line with VARGPT's way of asking questions (for example: Please help me draw a picture of the scenery in the Southern Hemisphere in June. Can you draw a picture of the scenery in the Southern Hemisphere in June for me? ...), it still failed to successfully output the picture. Finally, given its strong comprehension ability, ARMOR was able to successfully generate the text content corresponding to the question, and at the same time generate a picture that is highly relevant to the text.  

\subsection{WoHG Algorithm Explore}

In order to endow the model with generation capabilities while preserving its original abilities, we employed a full-scale fine-tuning method using our dataset and obtained the results shown in Table ~\ref{tab:single_train_all}.

\begin{table}[htbp]
    \centering
    \begin{tabular}{|c|c|c|c|c|c|}
        \hline
        Epoch & Text\_L. & Emb. & Visual\_L. & Adp. & Und. \\
        \hline
        10 & $\checkmark$ & $\times$ & $\checkmark$ & $\checkmark$ & 0.78 \\
        \hline
        10 & $\checkmark$ & $\checkmark$ & $\checkmark$ & $\checkmark$ & 0.72 \\
        \hline
        20 & $\checkmark$ & $\times$ & $\checkmark$ & $\checkmark$ & 0.64 \\
        \hline
        20 & $\checkmark$ & $\checkmark$ & $\checkmark$ & $\checkmark$ & 0.48 \\
        \hline
    \end{tabular}
    \caption{A Single Train Stage Result of Understanding Ability. Text\_L denotes the text output layer. Visual\_L denotes the visual output layer. Adp. denotes the added transformers adapter. Und. denote the percentage of the original InternVL2.5 capabilities mantained after training.
}
    \label{tab:single_train_all}
\end{table}

We conducted trials on different training schemes presented in the table.

Initially, we used a large dataset with interleaved text and images for training. During the training process, we tested the model's comprehension ability. The tests revealed that both the ``text output layer+embedding+ visual output layer+adapter" and ``text output layer+visual output layer+adapter" training schemes severely affected the inheritance of the original capabilities. When training text output layer, a large amount of text output information significantly impacts the output of text output layer, undermining the original comprehension ability.

Therefore, to minimize the impact on the output of text output layer, we explored a phased - training scheme. In the first phase, we used a small amount of data to enable the model to learn when to output text, images, and a combination of text and images.

In the first phase, to enable the model to acquire the ability to output image start token, we conducted training with a small amount of data. 

\begin{table}[htbp]
    \centering
    \begin{tabular}{|c|c|c|c|c|c|}
        \hline
        Stage & Text\_L. & Emb. & Visual\_L. & Adp. & Und. \\
        \hline
        Stage1 & $\checkmark$ & $\checkmark$ & $\checkmark$ & $\checkmark$ & 0.72 \\
        \hline
    \end{tabular}
    \caption{Train The First Stage with A Small Amount of Data.}
    \label{tab:train_stage1_und}
\end{table}

After testing, after the first-stage phased training, the model can distinguish output modes well while maintaining its original comprehension ability. This lays the foundation for subsequently endowing the model with generation capabilities. The result is shown in Table ~\ref{tab:train_stage1_und}.

In stage 2, there are two remaining schemes: the first is to train ``visual output layer+adapter", the second is to train embedding+visual output layer+adapter (which is the finally selected scheme)
When we only trained with the ``visual output layer+adapter" scheme, we found that the loss of the model remained at a relatively high level for a long time, and the quality of the generated images remained at a certain stage as follows Figure ~/ref{fig:combined}.
We believe that if the newly added embedding part of the model does not participate in the training, it may result in the failure of establishing a connection between the newly added image embedding of the model and the text embedding. Merely relying on the subsequent adapter and visual output layer cannot achieve the association between text information and image information.
Therefore, we finally decided to train the model with the ``embedding+visual output layer+adapter" scheme to realize the generation ability of the model.

\begin{figure}[H]
    \centering
    \begin{subfigure}[b]{0.15\textwidth}
        \includegraphics[width=\textwidth]{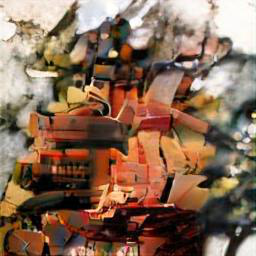}
        \label{fig:X_stage2_1}
    \end{subfigure}
    \begin{subfigure}[b]{0.15\textwidth}
        \includegraphics[width=\textwidth]{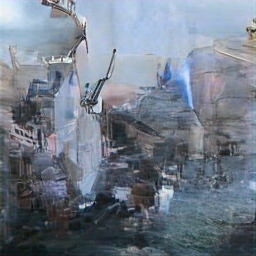}
        \label{fig:X_stage2_2}
    \end{subfigure}
    \caption{The Generated Images Trained with Visual Output Layer and Adapter.}
    \label{fig:combined}
\end{figure}